\documentclass{article}

\usepackage[square,numbers,compress]{natbib}
\PassOptionsToPackage{numbers, compress}{natbib}

\usepackage[final]{neurips_2023}

\usepackage{color,colortbl}

\usepackage[dvipsnames]{xcolor}
\definecolor{darkgreen}{rgb}{0,0.5,0}
\definecolor{azureblue}{rgb}{0,0.5,1}
\definecolor{darkgreen}{rgb}{1,0,0}
\definecolor{color1}{HTML}{006EB8}
\definecolor{color2}{HTML}{009B55}
\definecolor{color3}{HTML}{00A99A}
\definecolor{color4}{HTML}{3C8031}
\definecolor{color5}{HTML}{006795}
\definecolor{color6}{HTML}{00AEB3}
\definecolor{mygray}{gray}{0.93}
\definecolor{mygreen}{HTML}{3FBC9D}
\definecolor{arsenic}{rgb}{0.23, 0.27, 0.29}

\usepackage[utf8]{inputenc}
\usepackage[T1]{fontenc}
\usepackage{hyperref}
\hypersetup{
  colorlinks   = true,
  urlcolor     = color1,
  linkcolor    = color1,
  citecolor   = color1
}
\usepackage{url} 
\usepackage{booktabs}
\usepackage{amsfonts}
\usepackage{nicefrac}
\usepackage{microtype}
\usepackage{wrapfig}
\usepackage[nointegrals]{wasysym}
\usepackage{lipsum}  
\usepackage{times}
\usepackage{tikz}
\usepackage{latexsym}
\usepackage{microtype}
\usepackage{graphicx}
\usepackage{placeins}
\usepackage{subcaption}
\usepackage{bbm}
\usepackage{multirow}
\usepackage{enumitem}
\usepackage{tikz}
\usepackage[export]{adjustbox}
\usepackage{amsmath}
\usepackage{amssymb}
\usepackage{mathtools}
\usepackage{amsthm}
\usepackage{nccmath}
\usepackage{algorithm}
\usepackage{caption}
\usepackage{listings}
\usepackage[algo2e, ruled]{algorithm2e}
\SetKwComment{Comment}{$\triangleright$\ }{}
\SetKwProg{Init}{Initialize}{}{}

\usepackage{multirow}
\usepackage{pifont}
\newcommand{\cmark}{\ding{51}}
\newcommand{\xmark}{\ding{55}}
\newcommand{\greencmark}{{\color{mygreen}\cmark}}
\newcommand{\redxmark}{{\color{red}\xmark}}

\newcommand{\methodshort}[1]{\textsc{Ties-Merging}}
\newcommand{\method}[1]{\textsc{TrIm, Elect Sign \& Merge}}

\title{\methodshort{}: Resolving Interference When Merging Models}

\author{
    \textbf{Prateek Yadav}$^1$ \quad \textbf{Derek Tam}$^1$ \\ \quad
    \textbf{Leshem Choshen}$^{2,3}$ \quad \textbf{Colin Raffel}$^1$ \quad \textbf{Mohit Bansal}$^1$ 
  \\ 
  $^1$ University of North Carolina at Chapel Hill \quad $^2$ IBM Research \quad $^3$ MIT \\
  \texttt{leshem.choshen@ibm.com} \\
  \texttt{\{praty,dtredsox,craffel,mbansal\}@cs.unc.edu}
}

\begin{document}

\maketitle

\begin{abstract}

Transfer learning -- i.e., further fine-tuning a pre-trained model on a downstream task -- can confer significant advantages, including improved downstream performance, faster convergence, and better sample efficiency. These advantages have led to a proliferation of task-specific fine-tuned models, which typically can only perform a single task and do not benefit from one another. Recently, model merging techniques have emerged as a solution to combine multiple task-specific models into a single multitask model without performing additional training. However, existing merging methods often ignore the interference between parameters of different models, resulting in large performance drops when merging multiple models. In this paper, we demonstrate that prior merging techniques inadvertently lose valuable information due to two major sources of interference: (a) interference due to redundant parameter values and (b) disagreement on the sign of a given parameter's values across models. To address this, we propose our method, \method{} (\methodshort{}), which introduces three novel steps when merging models: (1) resetting parameters that only changed a small amount during fine-tuning, (2) resolving sign conflicts, and (3) merging only the parameters that are in alignment with the final agreed-upon sign.
We find that \methodshort{} outperforms several existing methods in diverse settings covering a range of modalities, domains, number of tasks, model sizes, architectures, and fine-tuning settings. We further analyze the impact of different types of interference on model parameters, and highlight the importance of resolving sign interference.\footnote{Our code is available at \href{https://github.com/prateeky2806/ties-merging}{https://github.com/prateeky2806/ties-merging}}

\end{abstract}

\setlength{\intextsep}{2pt}
\setlength{\columnsep}{2pt}

\section{Introduction}
\label{sec:introduction}
Pre-trained models (PTMs) have become widespread in many real-world applications~\cite{zhuang2020comprehensive,bommasani2021opportunities}. Using PTMs typically involves fine-tuning them to specialize on a specific task~\cite{shnarch2022label,devlin2018bert}, which can lead to improved performance with less task-specific labeled data. These benefits have resulted in the release of thousands of finetuned checkpoints~\cite{wolf2019huggingface} derived from popular PTMs such as ViT~\cite{dosovitskiy2021an} for vision and T5~\cite{raffel2019exploring} for language. However, having a separate fine-tuned model for each task has various drawbacks: (1) for each new application, a separate model has to be stored and deployed~\cite{fifty2021efficiently,zhang2021survey}, and (2) models trained in isolation cannot leverage information from related tasks to improve in-domain performance or out-of-domain generalization~\cite{sanh2022multitask,raffel2019exploring,tay2022ul2}.
Multitask learning~\cite{sanh2022multitask,colin2020exploring} could address these concerns but requires costly training and simultaneous access to all tasks~\cite{fifty2021efficiently}. Moreover, it can be complex and resource-intensive to determine how best to mix datasets to ensure that multitask training is beneficial for all tasks~\cite{pruksachatkun-etal-2020-intermediate,poth-etal-2021-pre,weller-etal-2022-use,phang2018sentence,albalak2023improving,fifty2021efficiently}.

\begin{figure}
    \centering
    \includegraphics[width=\linewidth]{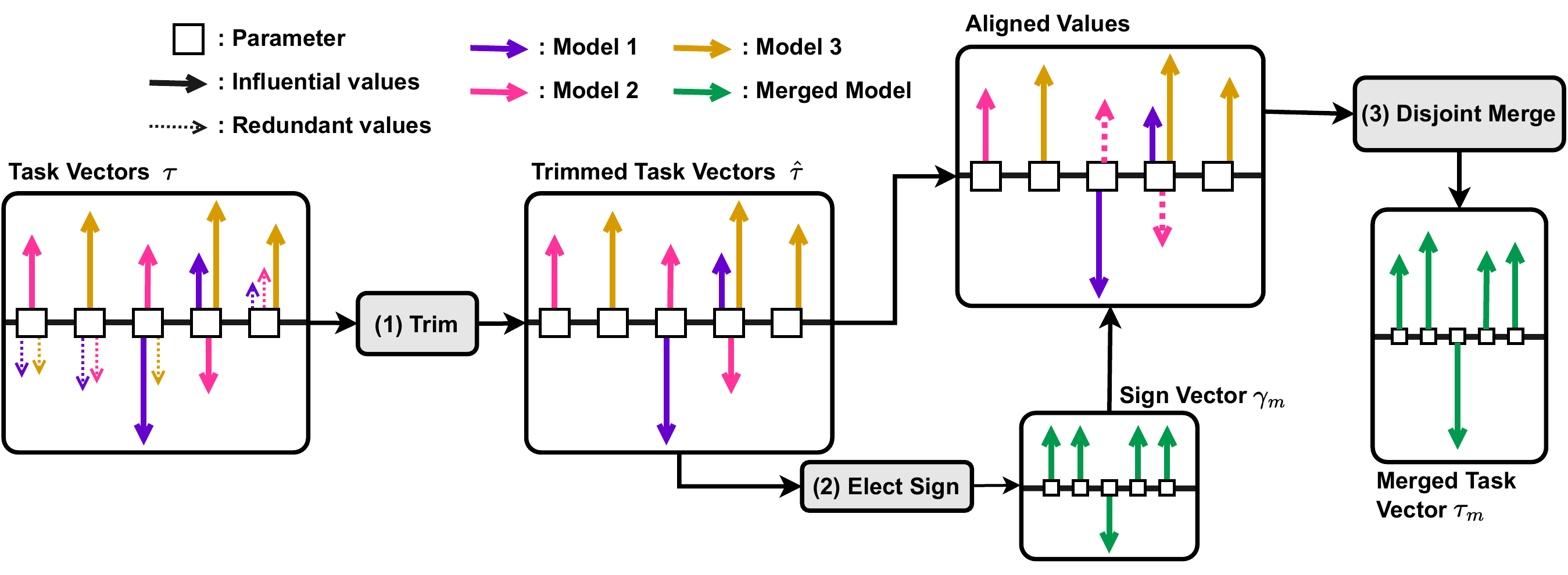}
    \captionsetup{type=figure}
    \caption{\label{fig:diagram_main} A depiction of the steps involved in \methodshort{}. We visualize each parameter in a model as a square. The arrows depict the update (task vector, $\tau$) to a parameter produced by fine-tuning on different tasks (coded by colors), with direction denoting sign and length denoting magnitude.
    We first \textit{trim} the task vector values based on their magnitude, then we \textit{elect} the sign for each parameter ($\gamma_m$, green vector containing $+1$ or $-1$) by resolving sign conflicts. Finally, we pick only the values that align with the elected sign and take their mean as the final parameter value.
}
\end{figure}

Recently, a growing body of research has focused on \textit{model merging} \cite{li2023deep}.
One application of merging involves combining multiple task-specific models into a single multitask model without performing additional training.
Previous works merge models by summing the individual model weights with different weighting schemes, either via a simple average ~\cite{choshen2022fusing,ilharco2022patching,wortsman2021robust}, via more sophisticated means that incorporate parameter importance~\cite{matena2021merging} or account for permutation invariances~\cite{ainsworth2022git,jin2023regmean,singh2020model,tatro2020optimizing,li2015convergent}. Combining fine-tuned models in this way can be seen as adding together \textit{task vectors}~\cite{ilharco2023editing} that are computed by subtracting the pre-trained model's parameter values from those of the fine-tuned model.

\setlength{\intextsep}{10pt}
\setlength{\columnsep}{10pt}
\begin{wrapfigure}{r}{0.45\textwidth}
\centering
\includegraphics[width=\linewidth]{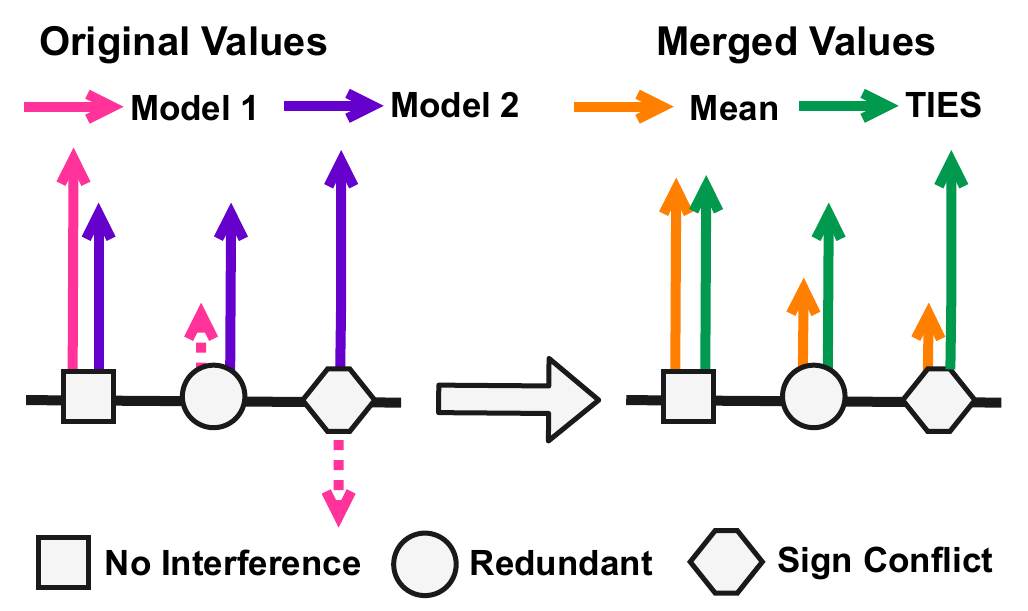}
\captionsetup{type=figure}
\caption{\label{fig:diagram_conflict} Different types of conflict and merged outputs produced by either averaging or \methodshort{}. The parameters causing interference are denoted by dotted arrows.
}
\end{wrapfigure}

While weighted averaging of model parameters has proven effective for merging, all of these methods ignore the possibility that values may interfere across models, thereby harming the performance of the merged model. In this paper, we first demonstrate that interference can stem from two major causes (see Fig. \ref{fig:diagram_conflict}), both of which can reduce parameter magnitudes in the merged model and eliminate subtle distinctions among values:
(1) \textsc{Interference from redundant parameters}: 
Previous studies on model pruning~\cite{hoefler2021sparsitysurvey,thimm1995evaluatingtopk} have shown that during fine-tuning, many model parameters can change over the course of fine-tuning~\cite{ruder2016overview} but only have a small impact on performance. However, when merging a parameter that is influential for one model but redundant (i.e.\ not influential) for other models, the influential value may be obscured by the redundant values, lowering the overall model performance ($\Circle$ in Fig. \ref{fig:diagram_conflict}). 
(2) \textsc{Interference from sign disagreement}: A given parameter might have a positive value for some models and a negative value for others.
Consequently, employing simple averaging might compromise the performance on both tasks ($\hexagon$ in Fig. \ref{fig:diagram_conflict}). 
In both of these situations, simply aggregating the values lead to interference that shrinks the parameter's value in the merged model. This interference between influential parameters might explain why the performance gap between the merged model and multitask-trained model increases as the number of models increases~\cite{jin2023regmean}.

To address these sources of interference, we propose \methodshort{} (\method{}) method, a method for merging models by combining task vectors that has three steps (visualized in Fig.~\ref{fig:diagram_main}):
First, we trim each task vector to retain only the influential parameter values by setting the redundant values in each task vector to zero (or, equivalently, resetting the fine-tuned parameter value back to the value from the pre-trained model). After this step, sign conflicts may still persist among influential parameter values, as visualized in Fig.~\ref{fig:imp_conflict}. 
Our second step therefore resolves the sign conflicts between different values and our last step only averages parameters whose sign agrees with the direction of the largest total movement across models.

We demonstrate the effectiveness of our proposed \methodshort{} method in various setups with:
(1) different modalities, including language and vision benchmarks, (2) distinct model sizes and families, such as T5-base and T5-large \cite{raffel2019exploring} as well as ViT-B/32 and ViT-L/14 \cite{dosovitskiy2021an}, (3) in-domain and out-of-domain tasks, (4) full finetuning or parameter-efficient finetuning, and (5) in the presence or absence of a validation set for setting merging hyperparameters. We show that \methodshort{} outperforms other merging methods, such as Task Arithmetic~\cite{ilharco2023editing}, RegMean~\cite{jin2023regmean}, Fisher Merging~\cite{matena2021merging}, and weight averaging~\cite{choshen2022fusing,wortsman2022model} across all these experimental settings. 
Notably, for in-domain evaluation, \methodshort{} outperforms the strongest baseline by an average of 2.3\% and 1.7\% absolute in NLP and vision settings (Table \ref{tab:main}), respectively. For out-of-domain generalization (Table \ref{tab:ood}), \methodshort{} outperforms the strongest baseline by $1.0\%$ and $4.4\%$ absolute for T5-base and T5-large models respectively. In Section \ref{sec:analysis}, we perform ablations on our method components and demonstrate the impact of interference on parameter values. Additionally, we showcase the increased advantage of \methodshort{} over task arithmetic~\cite{ilharco2023editing} as the number of tasks increases. Finally, we examine the importance of obtaining the correct sign vector. Our results and analysis establish \methodshort{} as a powerful and effective method for combining fine-tuned models into a single multi-task model.

\section{Related Work}
\label{sec:related_work}

\paragraph{Loss Landscape and Weight Interpolation.} While the loss function of a neural network is generally non-convex, recent work has demonstrated that the parameter values from different training runs can sometimes be interpolated without increasing the loss (i.e.\ they are \textit{mode-connected})~\cite{draxler2018essentially, freeman2016topology, garipov2018loss,jordan2023repair,gueta2023knowledge}. For example, \citet{frankle2020linear} showed that if a part of the optimization trajectory is shared between two neural networks then they can be interpolated without lowering accuracy. On the other hand, \citet{neyshabur2020being} showed that naively interpolating two neural networks with completely disjoint optimization trajectories can result in a catastrophic drop in their accuracies.  \citet{entezari2021role}  hypothesized that if we account for the permutation symmetry of neural networks, then all neural networks of a given architecture trained on the same dataset are linear mode connected. \citet{ainsworth2022git,singh2020model,wang2020federated} therefore used techniques based on finding permutations \cite{wang2020federated,ainsworth2022git} and optimal transport \cite{singh2020model} to better align neural networks trained from scratch so that they can be interpolated without increasing the loss.

\paragraph{Model Merging and Different Use Cases.}
Different fine-tuned models initialized from the same pre-trained model effectively share a part of the optimization trajectory, and can therefore often be merged without accounting for permutation symmetry~\cite{wortsman2022model,wortsman2021robust,ilharco2023editing,jin2023regmean}.
Therefore, merging fine-tuned models can improve performance on a single target task~\cite{izmailov2018averaging,gupta2020stochastic,wortsman2022model,choshen2022fusing}, improving out-of-domain generalization~\cite{jin2023regmean,ilharco2023editing,cha2021swad,arpit2021ensemble,Ram2022DiverseWA,rame2022modelrat}, creating a multitask models from different tasks~\cite{jin2023regmean,ilharco2023editing,li2022branch}, for federated learning~\cite{mcmahan2017communication,li2019convergence}, compression~\cite{li2023mc-smoe}, multimodal merging models~\cite{Sung2023AnEmpiricalSO}, continual learning~\cite{yadav2023codecl,yadav2023exssnet}, and other settings~\cite{li2022branch,don2022cold}. 
The range of applications has led to a proliferation of methods to improve beyond simple parameter averaging. 
\textit{RegMean}~\cite{jin2023regmean} proposed a closed-form solution for the merged model's parameters by solving a local linear regression problem for each individual linear layer in the model. However, this requires transmitting additional data statistics that are the same size as the model and requires additional inference steps to calculate them.
\textit{Fisher Merging}~\cite{matena2021merging} goes beyond simple averaging to identify the importance of individual parameters using Fisher Information Matrix~\cite{fisher1922mathematical,Amari1996fisher,kirkpatrick2017overcoming} and uses it to weigh the parameters in each model when merging. However, this shows little gains when merging multiple checkpoints and also requires computing gradients which has a high memory cost.
\textit{Task Arithmetic}~\cite{ilharco2023editing} presented a method for merging models by generating task vectors and performing arithmetic operations, such as addition, to obtain a multitask checkpoint. A concurrent work by~\citet{ortizjimenez2023tangent} provided theoretical insights on model merging based on the weight disentanglement property that arises during pretraining. They showed that finetuning models in their tangent space enhance this property, leading to better-merged models. Our method follows these past works on model merging but additionally takes into account the interference between different parameters during merging.

\section{Background and Motivation}
\label{sec:background}

\paragraph{Problem Setting}
Given a set of tasks $\{t_1, \hdots, t_n\}$ and a pre-trained model such as T5~\cite{raffel2019exploring} or ViT~\cite{dosovitskiy2021an}, we either finetune the entire model or employ a parameter-efficient finetuning (PEFT) method~\cite{liu2022tfew,hu2021lora}.
In both cases, we denote the trainable parameters as $\theta$, the initialization as $\theta_\textrm{init}$, and the finetuned parameters as $\theta_\textrm{ft}$. 
In this paper, we assume access to finetuned model parameters $\theta_\textrm{ft}$ for multiple tasks and devise a method to merge the weights of these models into a single multitask model proficient on both in-domain and out-of-domain datasets.
We follow~\citet{ilharco2023editing} and perform merging with task vectors. Specifically, for a task $t$, the task-vector $\tau_{t} \in \mathbb{R}^\textrm{d}$ is defined as $\tau_{t} = \theta_\textrm{ft}^t - \theta_\textrm{init}^t$. This operation allows us to focus on the changes that happen during the fine-tuning phase of each task-specific model and is equivalent to computing a weighted average of the models' weights with appropriate scaling.

\begin{figure}[t!]
  \centering
  \begin{minipage}[t]{0.49\linewidth}
    \centering
    \includegraphics[width=\linewidth]{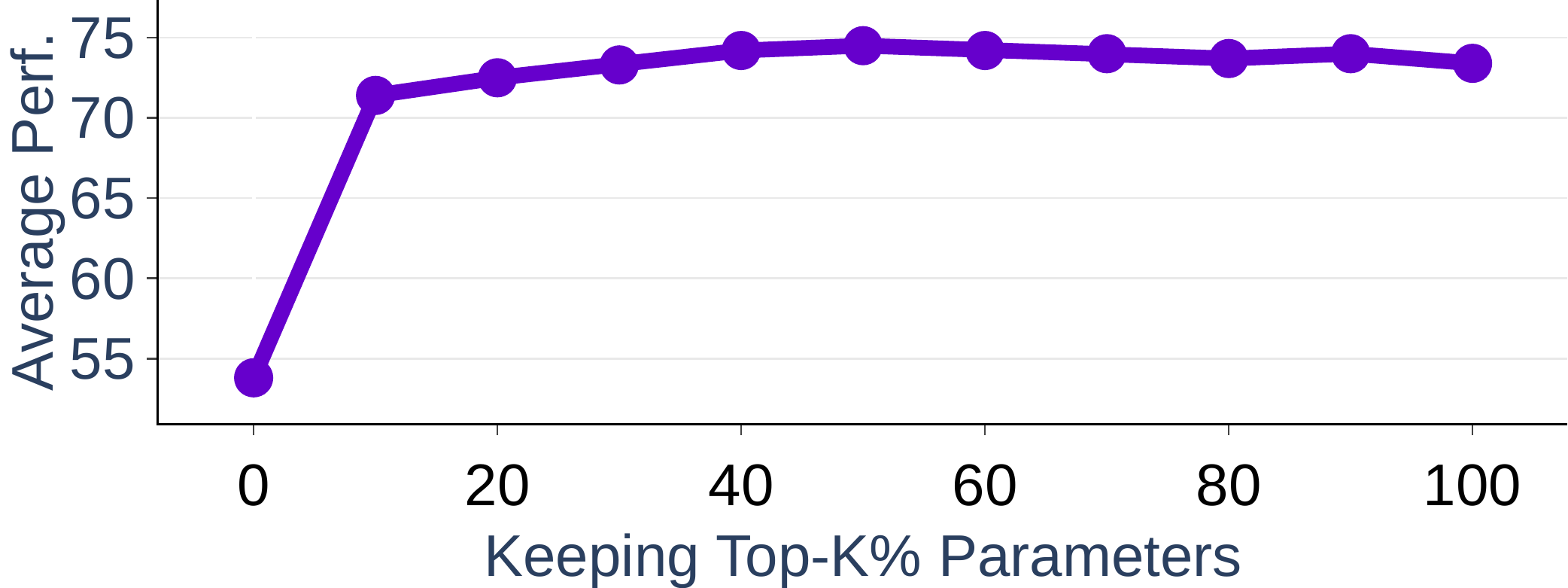}
    \captionsetup{type=figure}
    \caption{\label{fig:reset-bottomk} \textbf{Performance depends on a small fraction of high-magnitude parameters.} For each task vector, we keep only the largest - top-$k\%$ parameters and plot the average performance across eleven tasks. Keeping only the top-$20\%$ of the parameter does not degrade the performance.
    }
  \end{minipage}
  \hfill
  \begin{minipage}[t]{0.49\linewidth}
    \centering
    \includegraphics[width=\linewidth]{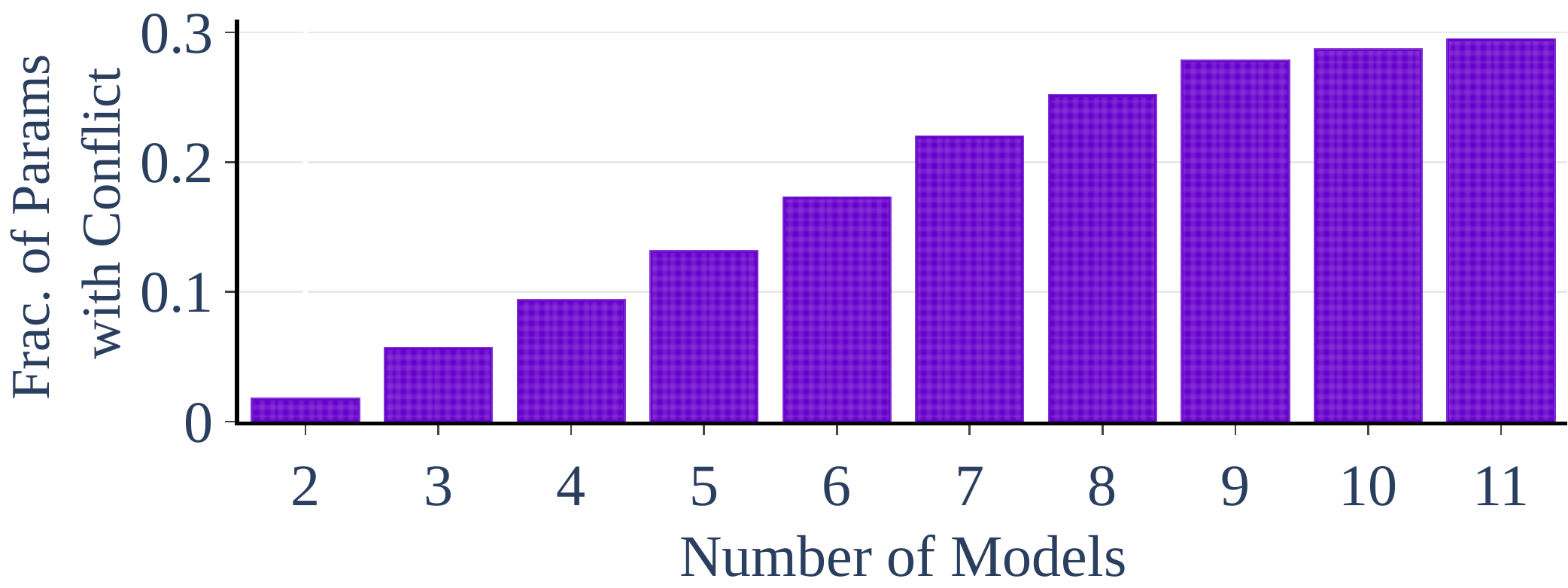}
    \captionsetup{type=figure}
    \caption{\label{fig:imp_conflict} \textbf{Sign conflicts occur even after trimming and increase with the number of models.}
    We plot the fraction of parameters that have a sign conflict after trimming versus the number of models being merged.
    }
  \end{minipage}
\end{figure}

\setlength{\intextsep}{2pt}
\setlength{\columnsep}{10pt}
\begin{wrapfigure}{r}{0.53\textwidth}
\begin{minipage}{\linewidth}
\begin{algorithm}[H]
\caption{\label{alg:merging} \methodshort{} Procedure.}
\DontPrintSemicolon
\KwIn{Fine-tuned models $\{\theta_t\}_{t=1}^n$, Initialization $\theta_\textrm{init}$, $k$, and $\lambda$.}
\KwOut{Merged Model $\theta_m$}
\ForAll{$t ~\textbf{in} 1,..., n$}{
    \Comment{\scriptsize Create task vectors.}
    $\tau_t = \theta_t - \theta_\textrm{init}$
    
    \Comment{\scriptsize Step 1: Trim redundant parameters.}
    $\hat{\tau}_t \leftarrow \text{keep\_topk\_reset\_rest\_to\_zero}(\tau_t, k)$

    $\hat{\gamma}_t \leftarrow sgn(\hat{\tau_t})$

    $\hat{\mu}_t \leftarrow |\hat{\tau_t}|$
}

\Comment{\scriptsize Step 2: Elect Final Signs.}

$\gamma_m = sgn(\sum_{t=1}^{n} \hat{\tau}_t)$

\Comment{\scriptsize Step 3: Disjoint Merge.}
\ForAll{$p ~\textbf{in} 1,..., d$}{
        $\mathcal{A}^p = {\{t \in [n] ~|~ \hat{\gamma}^p_t = \gamma_m^p\}}$
        
        $\tau^p_m = \frac{1}{|\mathcal{A}^p|}\sum_{t \in \mathcal{A}^p} \hat{\tau}_t^p$
    
    }

\Comment{\scriptsize Obtain merged checkpoint}

$\theta_m \leftarrow \theta_\textrm{init} + \lambda * \tau_m$

\Return{$\theta_m$} 
\end{algorithm}
\end{minipage}
\end{wrapfigure}

\paragraph{Redundancies in Model Parameters.}

First, we demonstrate that in a given task vector, many values are redundant (denoted by $\Circle$ in Fig.~\ref{fig:diagram_conflict}), and removing them does not affect the performance of the task.
Specifically, Fig.~\ref{fig:reset-bottomk} shows the average performance across eleven task-specific models when \textit{"trimming"} each task vector to retain only the top-$k\%$ largest-magnitude values and resetting the rest to their initial value (i.e.\ setting the corresponding value in the task vector to $0$). Fig.~\ref{fig:reset-bottomk} shows the average performance across varying values of $k$, demonstrating that keeping only the top-$20\%$ of values delivers comparable results to retaining all parameters. For additional details and the results on the T5 model, please refer to Appendix~\ref{sec:app_motivation}. This shows that many parameter changes introduced during fine-tuning are redundant. Hence, disregarding those values during merging could prevent interference with the influential parameters without compromising the task's performance.

\paragraph{Disagreement between Parameter Signs:} 
Different fine-tuned models might introduce opposing changes to a parameter in their task vectors, causing interference due to conflicting signs (denoted by $\hexagon$ in Fig.~\ref{fig:diagram_conflict}). 
Fig.~\ref{fig:imp_conflict} presents an analysis of the frequency of sign conflicts when merging varying numbers of models. We first trim the task vectors for eleven tasks by keeping only the top 20\% of influential parameters. Then, we plot the percentage of parameters that have a sign conflict as we increase the number of models to be merged from 2 to 11. Notably, sign conflicts occur even when merging only 2 models from different tasks or when merging multiple models from the same task (see Appendix Figure~\ref{fig:same_task_interference}), and the likelihood of a sign conflict increases with the number of models being merged. For additional details and the results on the T5 model, please refer to Appendix~\ref{sec:app_motivation}.

\section{\methodshort{}: \method{}}
\label{sec:method}

To address the aforementioned issues, we present \methodshort{} (\method{}), which aims to address the kinds of interference mentioned above before performing merging.

\subsection{Preliminaries}

A task vector $\tau_t \in \mathbb{R}^d$ represents a direction and the amount of movement required in the $d$-dimensional parameter space relative to the initialization that leads to a low loss region for the task $t$. 
Each entry in $\tau_t$ (corresponding to a particular parameter) can be thought of as an axis in the $d$-dimensional space. The sign of a parameter denotes the direction along this axis (positive or negative) that decreases the loss on task $t$.
Hence, a given task-vector $\tau_t$ can be decomposed into a \textit{sign vector} $\gamma_{t} \in \mathbb{R}^\textrm{d}$ and a \textit{magnitude vector} $\mu_{t} \in \mathbb{R}^\textrm{d}$ as $\tau_t = \gamma_t \odot \mu_t$, where $\odot$ is the elementwise product.
Formally, $\gamma_t = \textrm{sgn}(\tau_t)$, where $\textrm{sgn}(x) * |x| = x$ and returns a value of $+1$, $0$, or $-1$.
The magnitude vector $\mu_t$ is defined as $\mu_t = |\tau_t|$ and the value $\mu_t^i$ tells us the movement required in the $i$-th dimension from the initialization.

\subsection{Steps in \methodshort{}}
\label{sec:method_steps}

To merge multiple task-specific models $\{\theta_{t}\}_{t=1}^{n}$, we first create corresponding task vectors $\{\tau_{t}\}_{t=1}^{n}$. Given these task vectors, \methodshort{} method follows three steps in order to perform a merge  (see Fig.~\ref{fig:diagram_main} for a diagram and Algorithm~\ref{alg:merging}):

\begin{enumerate}[leftmargin=2em,labelsep=0.5em]
    
    \item \textbf{Trim:} For each task $t$, we trim the redundant parameters from the task vector $\tau_t$ to create $\hat{\tau}_t$ by keeping the top-$k\%$ values according to their magnitude and trimming the bottom $(100-k)\%$ of the redundant parameters by resetting them to 0. This can be decomposed further as $\hat{\tau}_t = \hat{\gamma}_t \odot \hat{\mu}_t$.
    
    \item \textbf{Elect:} Next, we create an aggregate elected sign vector $\gamma_m$ for the merged model that resolves the disagreements in the sign for each parameter $p$ across different models.
    To create the elected sign vector, we choose the sign with the highest total magnitude across all relevant models. For each parameter $p \in \{1,2, \hdots, d\}$, we separate the values $\{\hat{\tau}^p_t\}_{t=1}^{n}$ based on their sign ($+1$ or $-1$) and take their sum to calculate the total mass (i.e., total magnitude) in the positive and the negative direction. We then assign $\gamma_m^p$ as the sign with greater total movement. This can be efficiently computed using $\gamma_m^p = \textrm{sgn}(\sum_{t=1}^{n} \hat{\tau}_t^p)$.

    \item \textbf{Disjoint Merge:} Then, for each parameter $p$, we compute a \textit{disjoint mean} by only keeping the parameter values from the models whose signs are the same as the aggregated elected sign and calculate their mean. Formally, let $\mathcal{A}^p = {\{t \in [n] ~|~ \hat{\gamma}^p_t = \gamma_m^p\}}$, then $\tau^p_m = \frac{1}{|\mathcal{A}^p|}\sum_{t \in \mathcal{A}^p} \hat{\tau}_t^p$. Note that the disjoint mean always ignores the zero values. 

\end{enumerate}

Given the final merged task vector $\tau_m$, we scale it and add it to the initial parameter values to obtain the merged model parameters $\theta_m$ as $\theta_m = \theta_\textrm{init} + \lambda * \tau_m$, where $\lambda$ is a scaling hyperparameter (as used in past work~\citep{ilharco2023editing}).

\section{Experimental Setup}
\label{sec:exp_setup}

\paragraph{Baseline Methods.}
We compare \methodshort{} with four baseline merging methods: 
(1) \textbf{Simple Averaging} \cite{choshen2022fusing,wortsman2022model} calculates the element-wise mean of all the individual models and can be expressed as $\theta_m = \sum_{t=1}^{n} \theta_t / n$. 
(2) \textbf{Fisher Merging} \cite{matena2021merging}
uses a diagonal approximation of the Fisher Information Matrix $\hat{F}_t$ \cite{kirkpatrick2017overcoming,Amari1996fisher,fisher1922mathematical} to measure the importance of each parameter for task $t$, where $\hat{F}_t= \mathbb{E}_{x\sim D_t}\mathbb{E}_{y\sim p_{\theta_t}(y|x)} \nabla_{\theta_t} (\log p_{\theta_t}(y|x_t))^2$. The final merged model is obtained by reweighting each parameter in each fine-tuned model by the corresponding value in the model's approximate Fisher matrix as $\theta_{m}=\sum_{t=1}^{n} \hat{F}_t \theta_t / \sum_{t=1}^{n} \hat{F}_t$. 
(3) \textbf{RegMean} \cite{jin2023regmean} computes a closed-form solution to a least-squares regression problem that aims to minimize the distance between the merged model's activations and the individual models' activations as $\theta_m = (\sum_{t=1}^n X_t^TX_t)^{-1} \sum_{t=1}^n (X_t^T X_t \theta_t)$, where $X_t$ is the input activation of a given layer.
(4) \textbf{Task Artithmetic} \cite{ilharco2023editing} scales and then adds the task vectors to the initial model to produce the merged model as $\theta_m = \theta_\textrm{init} + \lambda * \sum_{t=1}^n \tau_t$.
In addition to these baselines, we present the performance of the individual \textbf{fine-tuned models} involved in the merging process as well as the performance of a \textbf{multi-task model} trained on the concatenation of all tasks' datasets. For more details on compute resources, dataset licenses, and the finetuning procedures, refer to Appendix~\ref{sec:app_compute},~\ref{sec:app_dataset},~and~\ref{sec:app_training_details}.

\paragraph{Merging in Absence of the Validation Set.}
Prior works \cite{ilharco2023editing,matena2021merging,wortsman2022model} on model merging assume access to a validation set, which is utilized to compute the Fisher matrix or tune hyper-parameters. To avoid the need for a validation set, RegMean \cite{jin2023regmean} proposed storing and transmitting inner product matrices of the training data for each task that are the same size as the original model. This can quickly become expensive for large models as the storage and transmission scale linearly with model size and the number of tasks.

To consider the setting where no validation set is available, we developed a generic recipe of \methodshort{} with fixed hyperparameters that could be applied in any setting without hyperparameter tuning on a validation set. The recipe keeps the top-$20\%$ parameters in the task vector resetting the rest to 0 and sets $\lambda = 1$. We chose this recipe based on results in the parameter-efficient fine-tuning (PEFT) setting, so we only apply it to the unseen settings of full model fine-tuning on ViT (vision) and T5 (language) models. We also compare \methodshort{} with the Task Arithmetic method without a validation set by utilizing the recommended value of $\lambda = 0.4$~\cite{ilharco2023editing}. For further details on how this recipe was created please refer to Appendix~\ref{sec:app_validation}.

\section{Main Results}
\label{sec:main_results}

Our main goal is to merge multiple task-specific models into a single multitask model that can perform well on both in-domain and out-of-domain scenarios. In this section, we evaluate the performance of \methodshort{} with other methods across multiple different experimental settings.

\begin{table*}[t!]
\centering
\resizebox{\linewidth}{!}{  
\begin{tabular}{lclllll}
\toprule

\textbf{Method ($\downarrow$)} & \multicolumn{1}{c}{\textbf{Validation}}  & \multicolumn{1}{c}{\textbf{PEFT}} & \multicolumn{4}{c}{\textbf{Full Finetuning}} \\
\cmidrule(lr){2-2} \cmidrule(lr){3-3} \cmidrule(lr){4-7}
\textbf{Model ($\rightarrow$)}&  & \textbf{(IA)$^3$} & \textbf{T5-Base} & \textbf{T5-Large} & \textbf{ViT-B/32} & \textbf{ViT-L/14}\\

\midrule
\rowcolor{gray!20} \textsc{Fine-tuned} & - & 71.4 & 82.8 & 88.8 & 90.5 & 94.2 \\
\rowcolor{gray!20} \textsc{Multitask} & - & 73.1 & 83.6 & 88.1  & 88.9 & 93.5 \\
\midrule
\textsc{Averaging} \cite{wortsman2022model,choshen2022fusing} & \redxmark & - & 65.9 & 59.6 & 65.8 & 79.6 \\
\textsc{Task Arithmetic} \cite{ilharco2023editing} & \redxmark & - & \textbf{73.2} & 73.5 & 60.4 & 83.3 \\
\textbf{\textsc{\methodshort{}}} & \redxmark & - & 69.7 [\textcolor{BrickRed}{-3.2}] & \textbf{74.4} [\textcolor{color2}{+0.9}] & \textbf{72.4} [\textcolor{color2}{+6.6}] & \textbf{86.0} [\textcolor{color2}{+2.7}] \\

\midrule
\textsc{Fisher Merging} \cite{matena2021merging} & \greencmark & 62.2 & 68.9  & 64.6 & 68.3 & 82.2 \\
\textsc{RegMean} \cite{jin2023regmean} & \greencmark & 58.0 & 71.2 & 73.2 & 71.8 & 83.7 \\
\textsc{Task Arithmetic} \cite{ilharco2023editing} & \greencmark & 63.9 & 73.2 & 73.3 & 70.1 & 84.5 \\
\textbf{\textsc{\methodshort{}}} & \greencmark & \textbf{66.4} [\textcolor{color2}{+2.5}] & \textbf{73.9} [\textcolor{color2}{+0.7}] & \textbf{76.9} [\textcolor{color2}{+3.6}] & \textbf{73.6} [\textcolor{color2}{+1.8}] & \textbf{86.0} [\textcolor{color2}{+1.5}] \\ 

\bottomrule
\end{tabular}
}
\captionsetup{type=table}
\caption{\label{tab:main} Comparing model merging methods across multiple fine-tuning settings and modalities (NLP and Vision) with and without the availability of a validation set.}
\end{table*}

\paragraph{Merging PEFT Models.}
Consider the setting where task vectors are computed based on parameters introduced during parameter-efficient fine-tuning. Specifically, we focus on (IA)$^3$ \cite{liu2022tfew}, a PEFT method that scales the base model activations with learned vectors. We follow \citet{liu2022tfew} and use T0-3B \cite{sanh2022multitask} as the base model and finetune (IA)$^3$ models on the train split of eleven datasets including sentence completion (COPA \citep{copa}, H-SWAG \citep{zellers2019hellaswag}, and Story Cloze \citep{sharma2018tackling} datasets), natural language inference (ANLI \citep{nie2019adversarial}, CB \citep{cb}, and RTE \citep{dagan2005pascal}), coreference resolution (WSC \citep{wsc} and Winogrande \citep{sakaguchi2020winogrande}), and word sense disambiguation (WiC \citep{pilehvar2018wic}). When fine-tuning (IA)$^3$ parameters added to the T0-3B model, we use prompt templates from the Public Pool of Prompts (P3 \citep{bach2022promptsource}) to convert each example in each dataset to a prompted text-to-text format where each label corresponds to a different string. For experiments with (IA)$^3$, for each dataset, we report the median score across all templates.

Table \ref{tab:main} using \methodshort{} to merge models trained with (IA)$^3$ exceeds the performance of all other merging methods -- with a validation set, \methodshort{} shows an average enhancement of $2.5\%$ across 11 tasks compared to the top baseline. For detailed results, refer to Appendix Table~\ref{tab:app_main_ia3}.

\paragraph{Merging Fully Finetuned Vision Models.}
For image classification, we adhere to the experimental setting from \citet{ilharco2023editing,ilharco2022patching}. We employ two variants of the CLIP model \cite{radford2021learning} with ViT-B/32 and ViT-L/14 models \cite{dosovitskiy2021an} as visual encoders. Subsequently, we finetune the visual encoder on the eight tasks derived from \citet{ilharco2022patching,ilharco2023editing,radford2021learning} while keeping the text encoder fixed. This setting considers a variety of classification domains such as remote sensing, traffic classification, and satellite imagery recognition. Specifically, we work with the following datasets: Cars \citep{cars}, DTD \citep{dtd}, EuroSAT \citep{eurosat}, GTSRB \citep{gtsrb}, MNIST \citep{lecun1998mnist}, RESISC45 \citep{cheng2017remote}, SUN397 \citep{sun397}, and SVHN \citep{svhn}.

Table~\ref{tab:main} shows that using \methodshort{} to merge fully fine-tuned ViT-B/32 and ViT-L/14 models leads to an average improvement of $1.8\%$ and $1.5\%$ over 8 tasks, given the availability of a validation set. In the absence of a validation set, \methodshort{} improves by $6.6\%$ and $2.7\%$ over other methods for ViT-B/32 and ViT-L/14, respectively. Notably, \methodshort{} without validation outperforms Task Arithmetic \cite{ilharco2023editing} with validation by $2.3\%$ and $1.5\%$ for ViT-B/32 and ViT-L/14. For more detailed results, refer to Appendix Table~\ref{tab:app_main_vitbase}~and~\ref{tab:app_main_vitlarge}.

\paragraph{Merging Fully Finetuned NLP Models.}

For the NLP domain, we use the T5-base and T5-large \cite{colin2020exploring} models, which are encoder-decoder transformers \cite{vaswani2017attention} pretrained via masked language modeling on a large text corpus. We finetune both T5-base and T5-large on seven tasks: question answering (QASC \cite{khot2020qasc}, WikiQA \cite{yang-etal-2015-wikiqa}, and QuaRTz \cite{tafjord2019quartz}), Paraphrase Identification (PAWS \cite{paws2019naacl}), Sentence Completion (Story Cloze \cite{sharma2018tackling}), and Coreference Resolution (Winogrande \cite{sakaguchi2020winogrande} and WSC~\cite{wsc}).

Table \ref{tab:main} shows that using \methodshort{} on T5-base and T5-large models with a validation set produces an improvement of $0.7\%$ and $3.6\%$ respectively over 7 tasks compared to the state-of-the-art. Moreover, for T5-large \methodshort{} without validation outperforms all baselines (even with a validation set) by $1.1\%$. For more detailed results, refer to Appendix Table~\ref{tab:app_main_t5base}~and~\ref{tab:app_main_t5large}.

\begin{figure}[t!]
  \centering
    \begin{minipage}[b]{0.49\linewidth}
        \centering
        \resizebox{\linewidth}{!}{  
        \begin{tabular}{lll}
        \toprule
        
        \textbf{Model}  &  \textbf{T5-Base}  &  \textbf{T5-Large} \\
        \midrule
        \textbf{Zeroshot}  &  31.1  & 27.6 \\
        \midrule
        \textbf{Simple Averaging} \cite{choshen2022fusing,wortsman2022model} & 31.7 & 30.4\\
        \textbf{Fisher} \cite{matena2021merging} & 33.8 & 32.0 \\
        \textbf{RegMean} \cite{jin2023regmean}  & 34.3 & 36.0 \\
        \textbf{Task Arithmetic} \cite{ilharco2023editing}  & 31.9 & 32.3 \\
        \textbf{\methodshort{}}  & \textbf{35.3} [\textcolor{color2}{+1.0}] & \textbf{40.4} [\textcolor{color2}{+4.4}] \\
        \bottomrule
        \end{tabular}
        }
        \captionsetup{type=table}
        \caption{\label{tab:ood}\textbf{\methodshort{} generalizes better.} Out of Distribution Generalization for T5-Base and T5-Large on six held-out tasks.}
  \end{minipage}
  \hfill
  \begin{minipage}[b]{0.49\linewidth}
    \centering
    \includegraphics[width=\linewidth]{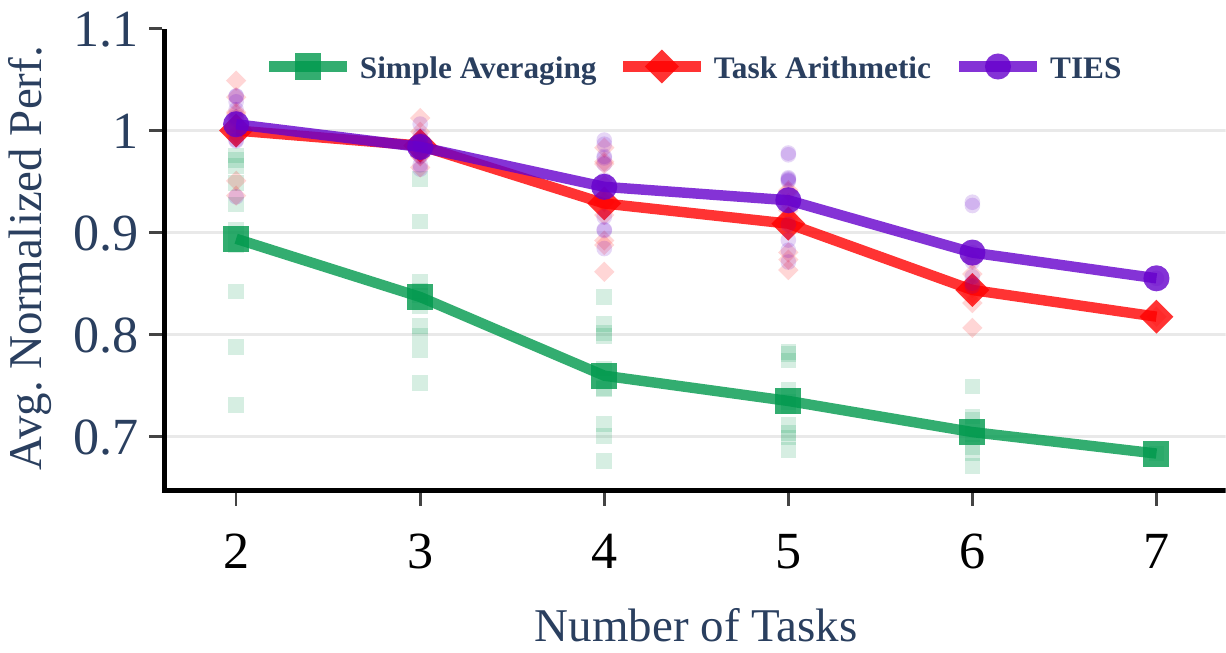}
    \captionsetup{type=figure}
    \caption{\label{fig:num_tasks} \textbf{\methodshort{} scales better.} Average performance when merging a different number of tasks.}
  \end{minipage}
\end{figure}

\paragraph{Out-of-Domain Generalization.}
In many use-cases, multitask models are used for their ability to generalize better to domain shift. Hence, we use the T5-base and T5-large models merged on the seven in-domain datasets from the previous experiments and evaluate them on six held-out datasets from T0 mixture~\cite{sanh2021multitask} to measure out-of-domain generalization. 
Specifically, we report the average performance over the following tasks and datasets: Cosmos QA \cite{huang2019cosmos}, Social IQA \cite{sap2019social}, and QuAIL \cite{quail_dataset} for question answering; WiC \cite{pilehvar2018wic} for word sense disambiguation; and COPA \cite{copa}, and H-SWAG \cite{zellers2019hellaswag} for sentence completion. Table \ref{tab:ood} shows that \methodshort{} outperforms the strongest baseline for both T5-base and T5-Large by $1.0\%$ and $4.4\%$ respectively, demonstrating better out-of-domain generalization. For more elaborate results please refer to Appendix~\ref{sec:app_per_task_results} and Table~\ref{tab:app_ood_t5base}~and~\ref{tab:app_ood_t5large}.

\paragraph{Merging Different Number of Tasks.}
We evaluate the performance of the merged model on the in-domain tasks as we vary the number of tasks being merged. In Fig.~\ref{fig:num_tasks}, we normalize the accuracy of each task by its fine-tuned model performance and report the average normalized accuracy on the in-domain tasks. We compare with the strongest baseline -- Task Arithmetic \cite{ilharco2023editing} -- as well as simple averaging \cite{wortsman2022model}. Each data point signifies merging a subset of the tasks, with the solid line representing the mean performance across multiple subsets. 
For similar results with the T5-base model, please refer to Appendix~\ref{sec:app_num_tasks} and Figure~\ref{fig:app_num_tasks_t5base}.

From Fig.~\ref{fig:num_tasks}, we observe the following: (1) As the number of merged tasks increases, the performance of all methods decreases. (2) When merging two tasks, both \methodshort{} and Task Arithmetic achieve an average normalized accuracy close to one, indicating negligible performance loss. In contrast, Simple Averaging suffers from a $10\%$ performance drop. (3) As the number of tasks increases, the merging performance of Task Arithmetic declines more rapidly than \methodshort{}.
This suggests that task interference is present when merging multiple tasks and that \methodshort{} is more effective at mitigating this issue.

\begin{figure}[t!]
  \centering
    \begin{minipage}{0.49\linewidth}
        \centering
        \resizebox{\linewidth}{!}{  
        \begin{tabular}{llll}
        \toprule
        
        	& \textbf{RTE} &	\textbf{MRPC} &	\textbf{WNLI} \\
         \midrule
        \textbf{Averaging} &	59.9 &	78.2 &	56.3\\
        \textbf{Fisher} &	65.7 &	81.4 &	52.1 \\
        \textbf{Ensembling} &	70.8 &	86.0 &	45.1 \\
        \midrule
        \textbf{Task Arithmetic} &	71.8 & 	86.0 &	\textbf{59.2} \\
        \textbf{\methodshort{}} &	\textbf{72.2} &	\textbf{86.8} &	58.8 \\
        \bottomrule
        \end{tabular}
        }
        \captionsetup{type=table}
        \caption{\label{tab:modelsoups}\textbf{Model soups experimental setup. TIES improves performance when merging checkpoints on the same tasks.} For each task, we merge 10 checkpoints from Huggingface hub and evaluate on the one task they were trained on.}
  \end{minipage}
  \hfill
  \begin{minipage}{0.49\linewidth}
    \centering
    \resizebox{\linewidth}{!}{  
    \begin{tabular}{llll}
    \toprule
    
    \textbf{Init Method} & \textbf{RTE} & \textbf{MRPC} & \textbf{WNLI} \\
    \midrule
    \textbf{PTM Init} & 66.4 & 81.8 & \textbf{56.3} \\
    \textbf{Average} & 75.8 & 86.5 & \textbf{56.3} \\
    \textbf{Task Arithmetic} & 78.3 & 86.2 & 50.7 \\
    \midrule
    \textbf{\methodshort{}} & \textbf{80.1} & \textbf{88.0} & 54.9 \\
     
    \bottomrule
    \end{tabular}
    }
    \captionsetup{type=table}
    \caption{\label{tab:fusing}\textbf{A TIES-merged model is a better initialization for finetuning.} For each task, we merge the checkpoints from the 7 other GLUE tasks and then finetune and evaluate on the selected task.}
  \end{minipage}
\end{figure}

\begin{figure}[t!]
  \centering
  \begin{subfigure}[b]{0.49\linewidth}
        \centering
        \includegraphics[width=\linewidth]{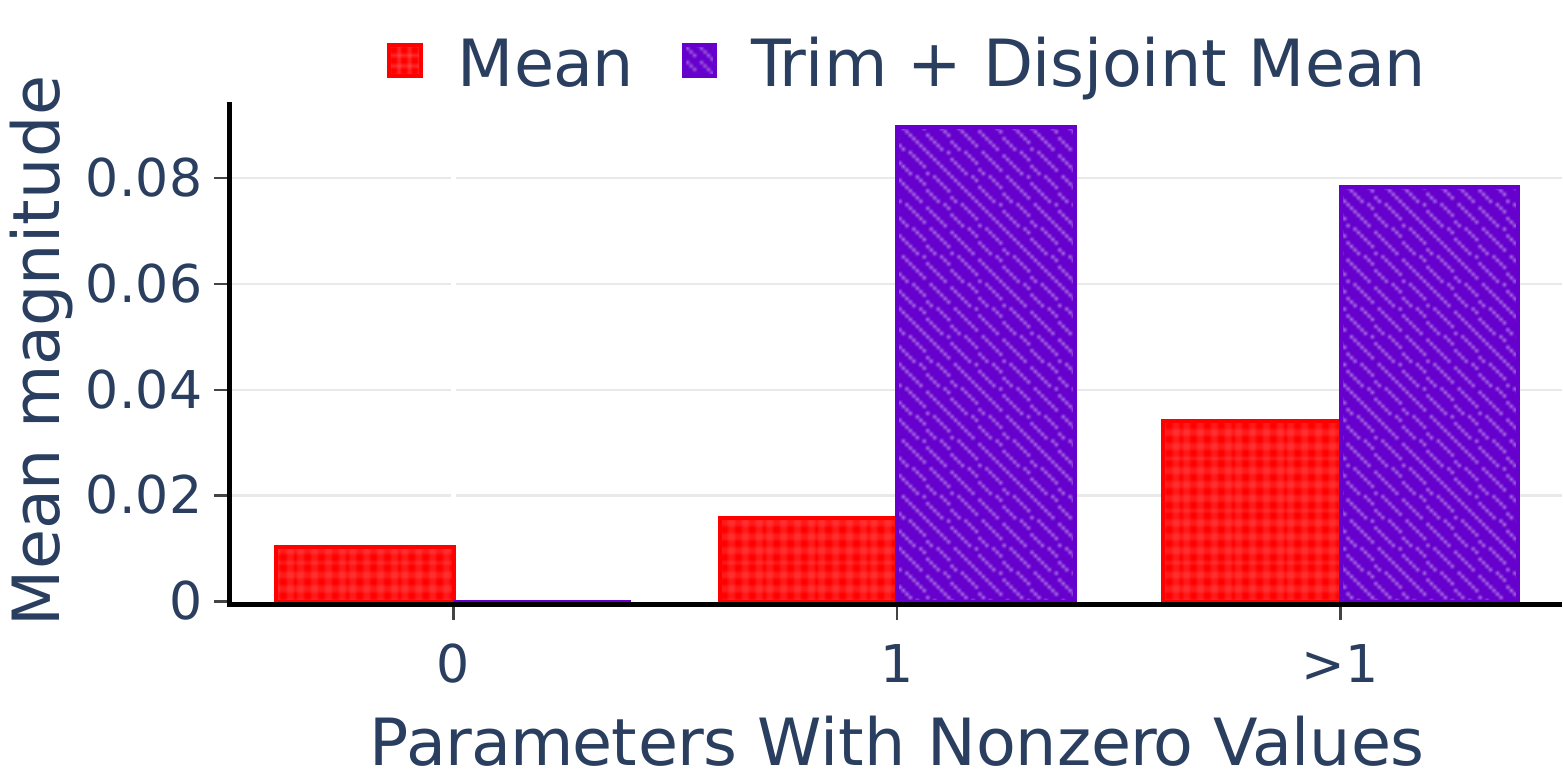}
        \caption{\label{fig:imp_noise} Redundant Parameter Interference.}
  \end{subfigure}
  \hfill
  \begin{subfigure}[b]{0.49\linewidth}
        \centering
        \includegraphics[width=\linewidth]{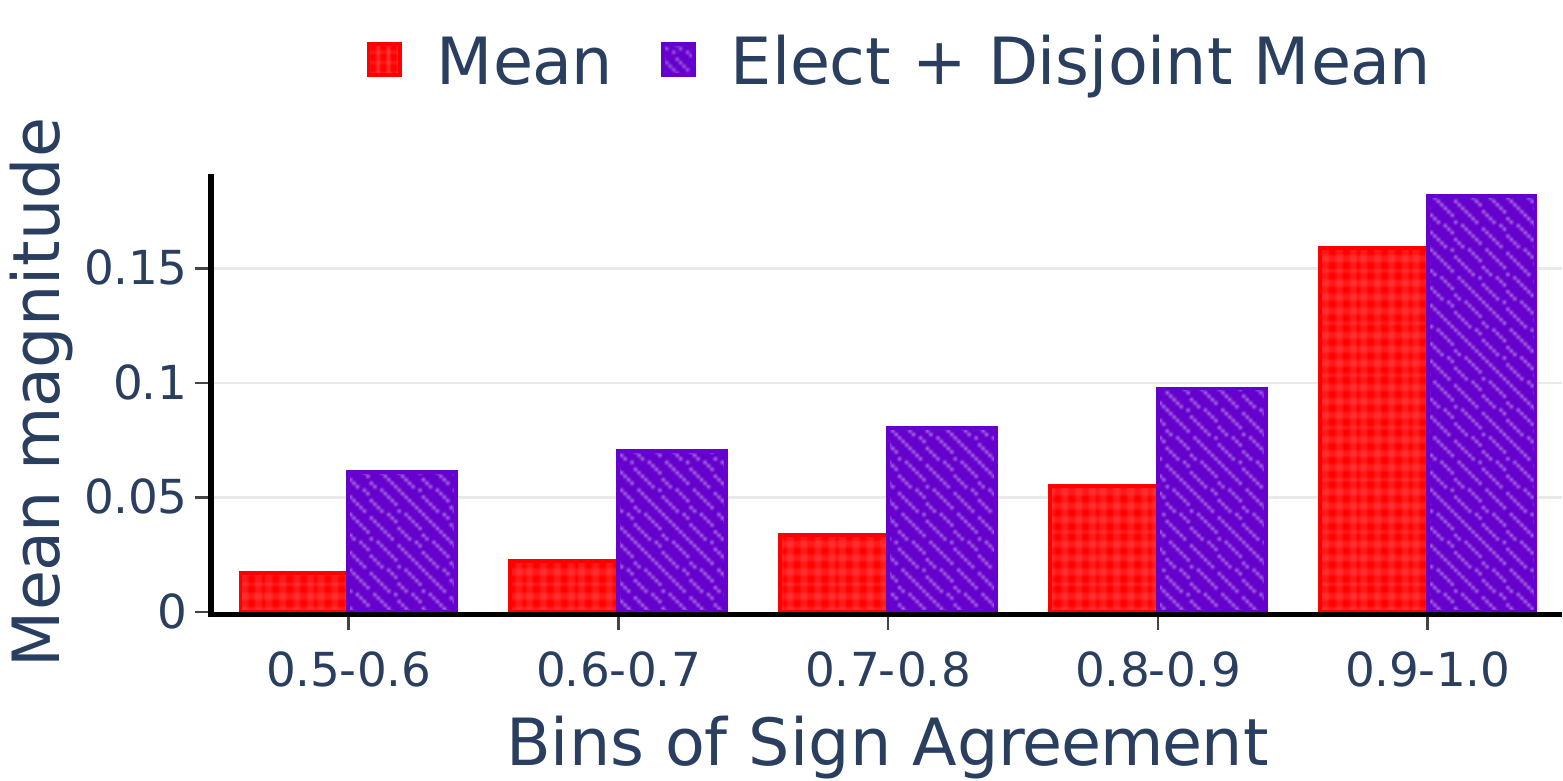}
        \caption{\label{fig:imp_sign} Sign Interference.}
    
  \end{subfigure}
  \caption{\label{fig:interference} \textbf{Trimming Parameters and Electing Signs prevents interference.}  Demonstration of parameter interference between different models and its impact on parameter values. The Standard Mean (red) shrinks magnitudes and does it more when there is less agreement on the sign (right) or the parameter is influential for multiple tasks (left).
  }
\end{figure}

\paragraph{Merging Checkpoints of the Same Task For Better Robustness}
We perform additional experiments to merge multiple checkpoints trained on the same task (as done in ModelSoups~\cite{wortsman2022model}) to see if it can improve robustness. Typically, ensembling is used to combine different models on the same task for better generalization. We use the experimental setting and the code from Fisher Merging~\cite{matena2021merging} to merge top-10 fine-tuned base sized BERT models from huggingface for RTE, MRPC, and WNLI datasets from GLUE.
From the results presented in Table~\ref{tab:modelsoups}, we observe that \methodshort{} works the best in all cases except WNLI, where it only slightly underperforms Task Vectors. Notably, \methodshort{} provides a dramatic boost over both Fisher Merging, averaging, and outperforms \emph{ensembling} in all cases. Moreover, in Appendix~\ref{sec:app_same_task_interference}, we show that interference exists even between differently finetuned checkpoints of the same tasks.

\paragraph{Merging Models for Better Initialization.}
Next, we perform experiments following the setting \cite{choshen2022fusing}, where we merge checkpoints from different tasks for a better initialization when fine-tuning on a downstream task. We take the finetuned \texttt{bert-base-uncased} checkpoints for 8 GLUE~\cite{wang2018glue} tasks (wnli, sst2, rte, qnli, mrpc, cola, mnli, qqp) from Huggingface~\cite{wolf2019huggingface}. We consider three of these GLUE tasks (RTE, MRPC, WNLI) as our downstream tasks. When fine-tuning on a particular downstream task (say RTE), we merge all the checkpoints from the other seven tasks together (apart from the chosen task). From Table~\ref{tab:fusing}, we find that \methodshort{} works well in this setting and outperforms all other merging methods by a significant margin (apart from Averaging for WNLI).

\section{Additional Results and Analysis}
\label{sec:analysis}

\subsection{Types of Interference and Their Effect on Merging}
\label{analysis:interference}

\paragraph{(a) Importance of Removing Redundant Parameters.}
To better disentangle the effect of redundant parameters on the resulting magnitude of merged parameters, we separate the parameters into three groups: redundant parameters (using a trimming threshold of $20\%$), parameters that are influential to exactly one model, and parameters that are influential to more than one model. We then compare the parameter values when they are directly merged versus when they are first trimmed and then (disjointly) merged without electing signs. Specifically, we only take the mean of non-trimmed values.
The results are shown for the PEFT setting in Fig.~\ref{fig:imp_noise}, which demonstrates that redundant parameters cause interference. Specifically, we find that when a parameter is not an influential parameter for any of the task-specific models, the mean value is low, and therefore may be considered noise.
However, when only one model sees the parameter as influential, the merged value can still be low since other models assign a small value to this parameter. The merged value is larger when more models see the parameter as influential. When trimming, we see this interference is mostly avoided, and the average size is mostly the same whether one or more models consider a parameter influential. This is because we remove the effect of noisy parameters that unnecessarily decrease the magnitude (see $\Circle$ in Fig.~\ref{fig:diagram_conflict}).
In the Appendix~\ref{sec:app_effect_interference}, we bring a more detailed view, including a comparison to applying \methodshort{} and show reducing interference encourages diversity in specific parameter values (std) together with the similarity of their influence (mean).

\paragraph{(b) Importance of Resolving Sign Interference.}
To quantify the impact of sign interference, we group the parameters by their \textit{sign agreement}. A value of $0.5$ indicates an equal number of positive and negative signs for a given parameter across different models, whereas $1$ implies all the parameters have the same sign. We compare the parameter values when those are merged, or when sign disagreement is first resolved by election and then they are (disjointly) merged.
The results in the PEFT setting are shown in Fig.~\ref{fig:imp_sign}, where we demonstrate that the \textsc{Elect} step preserves the relative parameter magnitudes to avoid sign interference. Specifically, we find that resolving signs increases the overall parameter magnitudes across different ranges of sign agreements. Parameters with low agreement tend to be smaller on average regardless of the interference.  One potential cause could be that the sign from noisy parameters pulls the average down, as seen in Fig.~\ref{fig:imp_noise}.
We show in Appendix~\ref{sec:app_effect_interference} that combining both methods indeed reduces some of the difference, but not all, suggesting that a high agreement is correlated with overall influential parameters.

\begin{figure}[t!]
  \centering
  \begin{minipage}{0.53\linewidth}
    \centering
    \includegraphics[width=\linewidth]{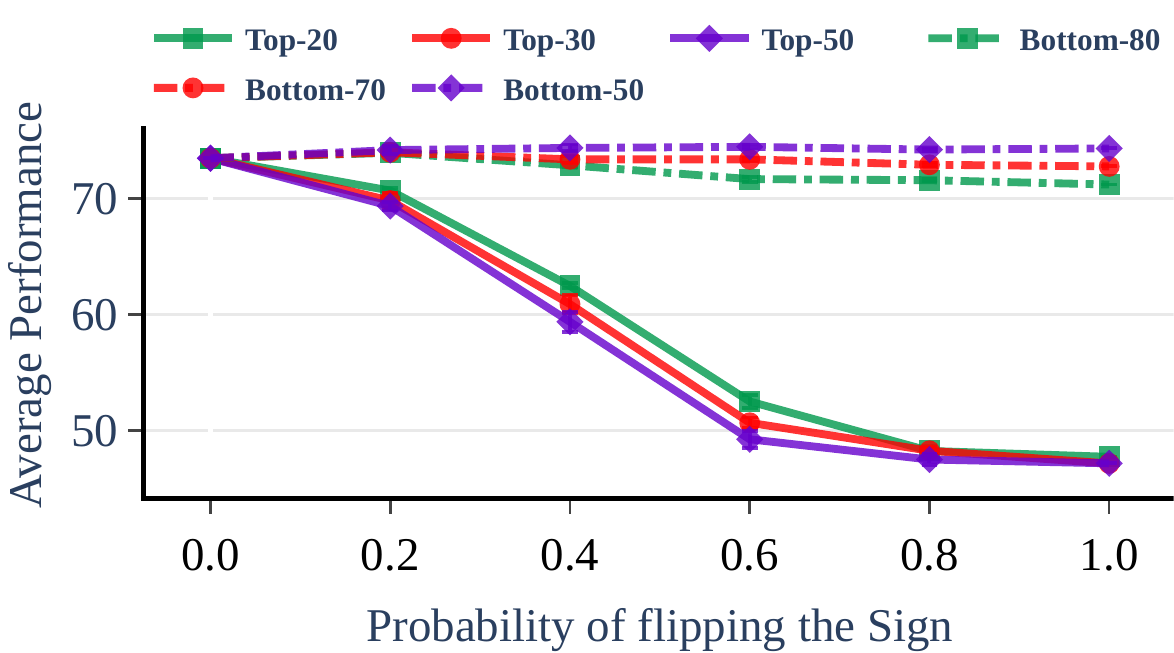}
    \captionsetup{type=figure}
    \caption{\label{fig:flip_signs} \textbf{Flipping the signs of high magnitude parameters leads to catastrophic performance drops.} Average Performance when flipping the directions of Top-$k\%$ and Bottom-$k\%$ parameters for each task. We report the results averaged over eleven (IA)$^3$ tasks.}
  \end{minipage}
  \hfill
  \begin{minipage}{0.43\linewidth}
        \centering
        \large
        \resizebox{\linewidth}{!}{  
        \begin{tabular}{ll}
        \toprule
        \textbf{Method} & \multicolumn{1}{l}{\textbf{Average}} \\
        \midrule
        \textbf{Fine-Tuned} & 71.4 \\
        \textbf{Multitask} & 73.1  \\
        \midrule
        \textbf{Averaging}~\cite{choshen2022fusing,wortsman2022model} & 58.0 \\
        \textbf{Task Vectors}~\cite{ilharco2023editing} & 63.9 \\
        \textbf{\methodshort{}} & \textbf{66.4} \\
        
        \midrule
        \textbf{\methodshort{} (Oracle Sign)} & \textbf{72.0} [\textcolor{color2}{+5.6}] \\
        
        \bottomrule
        \end{tabular}
        }
        \captionsetup{type=table}
        \caption{\label{tab:oracle} \textbf{\methodshort{} can perform close to multitask models if the signs can be estimated correctly.} 
        We use the signs from the multitask vector as the elected sign and perform merging and report the performance. 
        }
  \end{minipage}
\end{figure}

\subsection{Relevance of Signs of the Top-k\% Parameters}
\label{analysis:topk}

In this experiment, we work with the (IA)$^3$ models and aim to quantify the importance of the top-$k\%$ parameters and their directions on a task's performance. For each task vector $\tau_t$, we flip the direction of each of the top-$k\%$ parameters (by magnitude) with a probability $p$ to obtain $\Tilde{\tau}_t$. Flipping the direction is done by multiplying the parameters with $-1$. Then we add back this direction flipped $\Tilde{\tau}_t$ to the initialization to get $\Tilde{\theta}_t = \theta_\textrm{init} + \Tilde{\tau}_t$. Finally, we evaluate $\Tilde{\theta}_t$ and compute the average performance over all tasks $t$ for each value of $k$ and $p$. As a baseline, we also flip the directions of the bottom $(100-k)\%$ of the parameters. We report the results averaged over three independent runs.

In Fig.~\ref{fig:flip_signs}, we plot the average performance as a function of $p$, the probability of flipping the direction. A probability of $0$ means that the direction of none of the top-$k\%$ parameters is flipped and a value of $1$ means that the direction of all the top-$k\%$ parameters is flipped. For the top-$20/30\%$ of the parameters (solid lines), we observe that the performance monotonically decreases as we increase the probability of flipping the direction. In contrast, flipping the directions of the bottom-$80/70\%$ of the parameters (dashed lines) has little impact on the performance. These results establish the importance of having the right directions for the parameters with a high magnitude and show the catastrophic performance drop that happens with incorrect directions.

\subsection{Ablation of \methodshort{} Components}
\label{analysis:ablations}

\setlength{\intextsep}{3pt}
\setlength{\columnsep}{10pt}
\begin{wrapfigure}{r}{0.4\textwidth}
\centering
\resizebox{\linewidth}{!}{  
\begin{tabular}{lcc}
\toprule

\textbf{Method} & \textbf{T5-base} & \textbf{(IA)$^3$} \\
\midrule
\textbf{\textsc{\methodshort{}}} & \textbf{74.5} & \textbf{70.7}  \\
\midrule
\; $-$ \textsc{Trim} & 73.0 & 70.6 \\
\; $-$ \textsc{Elect} & 73.1 & 69.6 \\
\; $-$ \textsc{Disjoint Mean} & 72.6 & 67.5 \\
\; $-$ \textsc{Scale} & 72.0 & 65.5 \\

\bottomrule
\end{tabular}
}
\captionsetup{type=table}
\caption{\label{tab:ablation} Ablation on all the steps of \methodshort{}.}

\end{wrapfigure}

We perform ablations on the individual components of \methodshort{} to assess their importance. In Table~\ref{tab:ablation}, we start with \methodshort{} and remove one component at a time and report the performance on the validation set for full model merging (T5-base) and merging PEFT models ((IA)$^3$ on T03B). Removing elect while keeping the disjoint mean refers to taking the mean of values with signs $+1$ and $-1$ but not including the $0$ values of the trimmed task vectors in the mean. Removing disjoint mean but trimming and electing refers to taking the mean of the values with the elected signs and the $0$ for the trimmed values. Removing scaling means using $\lambda=1$.
Table~\ref{tab:ablation} shows that all components of the method are crucial for optimal performance. Specifically, scaling and the disjoint mean emerge as the most crucial, causing performance declines of $2.5\%$ and $1.9\%$ in T5-base, and $5.2\%$ and $3.2\%$ in (IA)$^3$, respectively.

\subsection{Importance of Estimating Correct Signs When Merging Models}
\label{analysis:estimating_sign}

Given the importance of sign vectors, we now aim to understand the performance that can be obtained by \methodshort{} if we can use the oracle sign vector from the multitask model. To test this, we train a multitask (IA)$^3$ model, $\theta_\textrm{mult}$, on the eleven tasks under consideration (as in \S~\ref{sec:main_results}). We then create the multitask vector $\tau_\textrm{mult}$ and the multitask sign vector $\gamma_\textrm{mult}$. Next, while merging models using \methodshort{}, we assume access to the oracle multitask-sign-vector $\gamma_\textrm{mult}$. 
Hence, we skip the conflict resolution step and directly set $\gamma_m = \gamma_\textrm{mult}$.
Surprisingly, from Table \ref{tab:oracle}, we observe that when merging tasks by using the oracle sign vector, we get a performance of $72\%$ compared to $73.1\%$ for the multitask trained model. Moreover, on average the merged model performs better task-specific models. This implies that if we can obtain the correction directions for the merged model, then we can closely bridge the gap to multitask models. In Appendix~\ref{sec:app_estimating_multitask_sign} and Table~\ref{tab:app_oracle}, we attempt to estimate the multitask-sign-vector by using limited data.

\section{Conclusion}
We introduced \methodshort{} to address interference when merging models. \methodshort{} trims low-magnitude changes in fine-tuned model's values and then resolves sign disagreements across the models being merged. We found experimentally that \methodshort{} enhances the performance of the merged multitask model across various settings and domains, despite being simple with fixed hyperparameters. Our study also sheds light on the impact of different types of interference on model parameters and emphasizes the importance of signs in the merging process. For some discussion on limitations and future directions please refer to Appendix~\ref{sec:limitation}.

\section*{Acknowledgements}

We thank Yi-Lin Sung, Shiyue Zhang, Archiki Prasad, and the reviewers for their valuable feedback on this paper. This work is supported by NSF-AI Engage Institute DRL211263, NSF-CAREER Award 1846185, DARPA MCS Grant N66001-19-2-4031, and NSF Grant 2145822. The views, opinions, and/or findings contained in this article are those of the authors and not of the funding agency.

\bibliography{references}

\begin{thebibliography}{91}
\providecommand{\natexlab}[1]{#1}
\providecommand{\url}[1]{\texttt{#1}}
\expandafter\ifx\csname urlstyle\endcsname\relax
  \providecommand{\doi}[1]{doi: #1}\else
  \providecommand{\doi}{doi: \begingroup \urlstyle{rm}\Url}\fi

\bibitem[Ainsworth et~al.(2022)Ainsworth, Hayase, and
  Srinivasa]{ainsworth2022git}
S.~K. Ainsworth, J.~Hayase, and S.~Srinivasa.
\newblock Git re-basin: Merging models modulo permutation symmetries, 2022.
\newblock \url{https://arxiv.org/abs/2209.04836}.

\bibitem[Albalak et~al.(2023)Albalak, Raffel, and Wang]{albalak2023improving}
A.~Albalak, C.~Raffel, and W.~Y. Wang.
\newblock Improving few-shot generalization by exploring and exploiting
  auxiliary data.
\newblock \emph{arXiv preprint arXiv:2302.00674}, 2023.

\bibitem[Amari(1996)]{Amari1996fisher}
S.~Amari.
\newblock Neural learning in structured parameter spaces - natural riemannian
  gradient.
\newblock In \emph{NIPS}, 1996.

\bibitem[Arpit et~al.(2021)Arpit, Wang, Zhou, and Xiong]{arpit2021ensemble}
D.~Arpit, H.~Wang, Y.~Zhou, and C.~Xiong.
\newblock Ensemble of averages: Improving model selection and boosting
  performance in domain generalization.
\newblock \emph{arXiv preprint arXiv:2110.10832}, 2021.

\bibitem[Bach et~al.(2022)Bach, Sanh, Yong, Webson, Raffel, Nayak, Sharma, Kim,
  Bari, F{\'e}vry, et~al.]{bach2022promptsource}
S.~H. Bach, V.~Sanh, Z.-X. Yong, A.~Webson, C.~Raffel, N.~V. Nayak, A.~Sharma,
  T.~Kim, M.~S. Bari, T.~F{\'e}vry, et~al.
\newblock {PromptSource}: An integrated development environment and repository
  for natural language prompts.
\newblock \emph{arXiv preprint arXiv:2202.01279}, 2022.

\bibitem[Bommasani et~al.(2021)Bommasani, Hudson, Adeli, Altman, Arora, von
  Arx, Bernstein, Bohg, Bosselut, Brunskill,
  et~al.]{bommasani2021opportunities}
R.~Bommasani, D.~A. Hudson, E.~Adeli, R.~Altman, S.~Arora, S.~von Arx, M.~S.
  Bernstein, J.~Bohg, A.~Bosselut, E.~Brunskill, et~al.
\newblock On the opportunities and risks of foundation models, 2021.
\newblock \url{https://arxiv.org/abs/2108.07258}.

\bibitem[Cha et~al.(2021)Cha, Chun, Lee, Cho, Park, Lee, and Park]{cha2021swad}
J.~Cha, S.~Chun, K.~Lee, H.-C. Cho, S.~Park, Y.~Lee, and S.~Park.
\newblock Swad: Domain generalization by seeking flat minima.
\newblock \emph{Advances in Neural Information Processing Systems},
  34:\penalty0 22405--22418, 2021.

\bibitem[Cheng et~al.(2017)Cheng, Han, and Lu]{cheng2017remote}
G.~Cheng, J.~Han, and X.~Lu.
\newblock Remote sensing image scene classification: Benchmark and state of the
  art.
\newblock \emph{Proceedings of the Institute of Electrical and Electronics
  Engineers (IEEE)}, 2017.
\newblock \url{https://ieeexplore.ieee.org/abstract/document/7891544}.

\bibitem[Choshen et~al.(2022)Choshen, Venezian, Slonim, and
  Katz]{choshen2022fusing}
L.~Choshen, E.~Venezian, N.~Slonim, and Y.~Katz.
\newblock Fusing finetuned models for better pretraining, 2022.
\newblock \url{https://arxiv.org/abs/2204.03044}.

\bibitem[Cimpoi et~al.(2014)Cimpoi, Maji, Kokkinos, Mohamed, and Vedaldi]{dtd}
M.~Cimpoi, S.~Maji, I.~Kokkinos, S.~Mohamed, and A.~Vedaldi.
\newblock Describing textures in the wild.
\newblock In \emph{Conference on Computer Vision and Pattern Recognition
  (CVPR)}, 2014.
\newblock
  \url{https://openaccess.thecvf.com/content_cvpr_2014/html/Cimpoi_Describing_Textures_in_2014_CVPR_paper.html}.

\bibitem[Dagan et~al.(2005)Dagan, Glickman, and Magnini]{dagan2005pascal}
I.~Dagan, O.~Glickman, and B.~Magnini.
\newblock The pascal recognising textual entailment challenge.
\newblock In \emph{Machine Learning Challenges Workshop}, 2005.
\newblock \url{https://link.springer.com/chapter/10.1007/11736790_9}.

\bibitem[Devlin et~al.(2018)Devlin, Chang, Lee, and Toutanova]{devlin2018bert}
J.~Devlin, M.-W. Chang, K.~Lee, and K.~Toutanova.
\newblock {BERT}: Pre-training of deep bidirectional transformers for language
  understanding.
\newblock \emph{arXiv preprint arXiv:1810.04805}, 2018.

\bibitem[Don-Yehiya et~al.(2022)Don-Yehiya, Venezian, Raffel, Slonim, Katz, and
  Choshen]{don2022cold}
S.~Don-Yehiya, E.~Venezian, C.~Raffel, N.~Slonim, Y.~Katz, and L.~Choshen.
\newblock Cold fusion: Collaborative descent for distributed multitask
  finetuning, 2022.
\newblock \url{https://arxiv.org/abs/2212.01378}.

\bibitem[Dosovitskiy et~al.(2021)Dosovitskiy, Beyer, Kolesnikov, Weissenborn,
  Zhai, Unterthiner, Dehghani, Minderer, Heigold, Gelly, Uszkoreit, and
  Houlsby]{dosovitskiy2021an}
A.~Dosovitskiy, L.~Beyer, A.~Kolesnikov, D.~Weissenborn, X.~Zhai,
  T.~Unterthiner, M.~Dehghani, M.~Minderer, G.~Heigold, S.~Gelly, J.~Uszkoreit,
  and N.~Houlsby.
\newblock An image is worth 16x16 words: Transformers for image recognition at
  scale.
\newblock In \emph{International Conference on Learning Representations
  (ICLR)}, 2021.
\newblock \url{https://openreview.net/forum?id=YicbFdNTTy}.

\bibitem[Draxler et~al.(2018)Draxler, Veschgini, Salmhofer, and
  Hamprecht]{draxler2018essentially}
F.~Draxler, K.~Veschgini, M.~Salmhofer, and F.~Hamprecht.
\newblock Essentially no barriers in neural network energy landscape.
\newblock In \emph{International Conference on Machine Learning (ICML)}, 2018.
\newblock \url{https://arxiv.org/abs/1803.00885}.

\bibitem[Entezari et~al.(2021)Entezari, Sedghi, Saukh, and
  Neyshabur]{entezari2021role}
R.~Entezari, H.~Sedghi, O.~Saukh, and B.~Neyshabur.
\newblock The role of permutation invariance in linear mode connectivity of
  neural networks.
\newblock \emph{arXiv preprint arXiv:2110.06296}, 2021.

\bibitem[Fifty et~al.(2021)Fifty, Amid, Zhao, Yu, Anil, and
  Finn]{fifty2021efficiently}
C.~Fifty, E.~Amid, Z.~Zhao, T.~Yu, R.~Anil, and C.~Finn.
\newblock Efficiently identifying task groupings for multi-task learning.
\newblock \emph{Advances in Neural Information Processing Systems},
  34:\penalty0 27503--27516, 2021.

\bibitem[Fisher(1922)]{fisher1922mathematical}
R.~A. Fisher.
\newblock On the mathematical foundations of theoretical statistics.
\newblock \emph{Philosophical transactions of the Royal Society of London.
  Series A, containing papers of a mathematical or physical character},
  222\penalty0 (594-604):\penalty0 309--368, 1922.

\bibitem[Frankle et~al.(2020)Frankle, Dziugaite, Roy, and
  Carbin]{frankle2020linear}
J.~Frankle, G.~K. Dziugaite, D.~Roy, and M.~Carbin.
\newblock Linear mode connectivity and the lottery ticket hypothesis.
\newblock In \emph{International Conference on Machine Learning (ICML)}, 2020.
\newblock \url{https://proceedings.mlr.press/v119/frankle20a.html}.

\bibitem[Freeman and Bruna(2016)]{freeman2016topology}
C.~D. Freeman and J.~Bruna.
\newblock Topology and geometry of half-rectified network optimization.
\newblock \emph{arXiv preprint arXiv:1611.01540}, 2016.

\bibitem[Garipov et~al.(2018)Garipov, Izmailov, Podoprikhin, Vetrov, and
  Wilson]{garipov2018loss}
T.~Garipov, P.~Izmailov, D.~Podoprikhin, D.~Vetrov, and A.~G. Wilson.
\newblock Loss surfaces, mode connectivity, and fast ensembling of dnns.
\newblock In \emph{Advances in Neural Information Processing Systems
  (NeurIPS)}, 2018.
\newblock \url{https://arxiv.org/abs/1802.10026}.

\bibitem[Gueta et~al.(2023)Gueta, Venezian, Raffel, Slonim, Katz, and
  Choshen]{gueta2023knowledge}
A.~Gueta, E.~Venezian, C.~Raffel, N.~Slonim, Y.~Katz, and L.~Choshen.
\newblock Knowledge is a region in weight space for fine-tuned language models.
\newblock \emph{arXiv preprint arXiv:2302.04863}, 2023.

\bibitem[Gupta et~al.(2020)Gupta, Serrano, and DeCoste]{gupta2020stochastic}
V.~Gupta, S.~A. Serrano, and D.~DeCoste.
\newblock Stochastic weight averaging in parallel: Large-batch training that
  generalizes well.
\newblock \emph{International Conference on Learning Representations}, 2020.

\bibitem[Helber et~al.(2019)Helber, Bischke, Dengel, and Borth]{eurosat}
P.~Helber, B.~Bischke, A.~Dengel, and D.~Borth.
\newblock Eurosat: A novel dataset and deep learning benchmark for land use and
  land cover classification.
\newblock \emph{Journal of Selected Topics in Applied Earth Observations and
  Remote Sensing}, 2019.
\newblock \url{https://arxiv.org/abs/1709.00029}.

\bibitem[Hoefler et~al.(2021)Hoefler, Alistarh, Ben-Nun, Dryden, and
  Peste]{hoefler2021sparsitysurvey}
T.~Hoefler, D.~Alistarh, T.~Ben-Nun, N.~Dryden, and A.~Peste.
\newblock Sparsity in deep learning: Pruning and growth for efficient inference
  and training in neural networks.
\newblock \emph{The Journal of Machine Learning Research}, 22\penalty0
  (1):\penalty0 10882--11005, 2021.

\bibitem[Hu et~al.(2021)Hu, Shen, Wallis, Allen-Zhu, Li, Wang, and
  Chen]{hu2021lora}
E.~J. Hu, Y.~Shen, P.~Wallis, Z.~Allen-Zhu, Y.~Li, S.~Wang, and W.~Chen.
\newblock {LoRA}: Low-rank adaptation of large language models.
\newblock \emph{ArXiv}, abs/2106.09685, 2021.

\bibitem[Huang et~al.(2019)Huang, Le~Bras, Bhagavatula, and
  Choi]{huang2019cosmos}
L.~Huang, R.~Le~Bras, C.~Bhagavatula, and Y.~Choi.
\newblock Cosmos qa: Machine reading comprehension with contextual commonsense
  reasoning.
\newblock In \emph{Proceedings of the 2019 Conference on Empirical Methods in
  Natural Language Processing and the 9th International Joint Conference on
  Natural Language Processing (EMNLP-IJCNLP)}, pages 2391--2401, 2019.

\bibitem[Ilharco et~al.(2022)Ilharco, Wortsman, Gadre, Song, Hajishirzi,
  Kornblith, Farhadi, and Schmidt]{ilharco2022patching}
G.~Ilharco, M.~Wortsman, S.~Y. Gadre, S.~Song, H.~Hajishirzi, S.~Kornblith,
  A.~Farhadi, and L.~Schmidt.
\newblock Patching open-vocabulary models by interpolating weights.
\newblock In \emph{Advances in Neural Information Processing Systems
  (NeurIPS)}, 2022.
\newblock \url{https://arXiv.org/abs/2208.05592}.

\bibitem[Ilharco et~al.(2023)Ilharco, Ribeiro, Wortsman, Schmidt, Hajishirzi,
  and Farhadi]{ilharco2023editing}
G.~Ilharco, M.~T. Ribeiro, M.~Wortsman, L.~Schmidt, H.~Hajishirzi, and
  A.~Farhadi.
\newblock Editing models with task arithmetic.
\newblock In \emph{The Eleventh International Conference on Learning
  Representations}, 2023.
\newblock URL \url{https://openreview.net/forum?id=6t0Kwf8-jrj}.

\bibitem[Izmailov et~al.(2018)Izmailov, Podoprikhin, Garipov, Vetrov, and
  Wilson]{izmailov2018averaging}
P.~Izmailov, D.~Podoprikhin, T.~Garipov, D.~Vetrov, and A.~G. Wilson.
\newblock Averaging weights leads to wider optima and better generalization.
\newblock In \emph{Conference on Uncertainty in Artificial Intelligence (UAI)},
  2018.
\newblock \url{https://arxiv.org/abs/1803.05407}.

\bibitem[Jin et~al.(2023)Jin, Ren, Preotiuc-Pietro, and Cheng]{jin2023regmean}
X.~Jin, X.~Ren, D.~Preotiuc-Pietro, and P.~Cheng.
\newblock Dataless knowledge fusion by merging weights of language models.
\newblock In \emph{The Eleventh International Conference on Learning
  Representations}, 2023.
\newblock URL \url{https://openreview.net/forum?id=FCnohuR6AnM}.

\bibitem[Jordan et~al.(2023)Jordan, Sedghi, Saukh, Entezari, and
  Neyshabur]{jordan2023repair}
K.~Jordan, H.~Sedghi, O.~Saukh, R.~Entezari, and B.~Neyshabur.
\newblock {REPAIR}: {RE}normalizing permuted activations for interpolation
  repair.
\newblock In \emph{The Eleventh International Conference on Learning
  Representations}, 2023.
\newblock URL \url{https://openreview.net/forum?id=gU5sJ6ZggcX}.

\bibitem[Khot et~al.(2020)Khot, Clark, Guerquin, Jansen, and
  Sabharwal]{khot2020qasc}
T.~Khot, P.~Clark, M.~Guerquin, P.~Jansen, and A.~Sabharwal.
\newblock Qasc: A dataset for question answering via sentence composition.
\newblock In \emph{Proceedings of the AAAI Conference on Artificial
  Intelligence}, volume~34, pages 8082--8090, 2020.

\bibitem[Kirkpatrick et~al.(2017)Kirkpatrick, Pascanu, Rabinowitz, Veness,
  Desjardins, Rusu, Milan, Quan, Ramalho, Grabska-Barwinska,
  et~al.]{kirkpatrick2017overcoming}
J.~Kirkpatrick, R.~Pascanu, N.~Rabinowitz, J.~Veness, G.~Desjardins, A.~A.
  Rusu, K.~Milan, J.~Quan, T.~Ramalho, A.~Grabska-Barwinska, et~al.
\newblock Overcoming catastrophic forgetting in neural networks.
\newblock \emph{Proceedings of the National Academy of Sciences (PNAS)}, 2017.
\newblock \url{https://arxiv.org/abs/1612.00796}.

\bibitem[Krause et~al.(2013)Krause, Stark, Deng, and Fei-Fei]{cars}
J.~Krause, M.~Stark, J.~Deng, and L.~Fei-Fei.
\newblock 3d object representations for fine-grained categorization.
\newblock In \emph{International Conference on Computer Vision Workshops
  (ICML)}, 2013.
\newblock
  \url{https://www.cv-foundation.org/openaccess/content_iccv_workshops_2013/W19/html/Krause_3D_Object_Representations_2013_ICCV_paper.html}.

\bibitem[LeCun(1998)]{lecun1998mnist}
Y.~LeCun.
\newblock The mnist database of handwritten digits, 1998.
\newblock \url{http://yann.lecun.com/exdb/mnist/}.

\bibitem[Levesque et~al.(2012)Levesque, Davis, and Morgenstern]{wsc}
H.~Levesque, E.~Davis, and L.~Morgenstern.
\newblock The winograd schema challenge.
\newblock \emph{Thirteenth International Conference on the Principles of
  Knowledge Representation and Reasoning}, 2012.

\bibitem[Li et~al.(2022)Li, Gururangan, Dettmers, Lewis, Althoff, Smith, and
  Zettlemoyer]{li2022branch}
M.~Li, S.~Gururangan, T.~Dettmers, M.~Lewis, T.~Althoff, N.~A. Smith, and
  L.~Zettlemoyer.
\newblock Branch-train-merge: Embarrassingly parallel training of expert
  language models, 2022.
\newblock \url{https://arxiv.org/abs/2208.03306}.

\bibitem[Li et~al.(2023{\natexlab{a}})Li, Zhang, Yadav, Sung, Cheng, Bansal,
  and Chen]{li2023mc-smoe}
P.~Li, Z.~Zhang, P.~Yadav, Y.-L. Sung, Y.~Cheng, M.~Bansal, and T.~Chen.
\newblock Merge, then compress: Demystify efficient smoe with hints from its
  routing policy, 2023{\natexlab{a}}.

\bibitem[Li et~al.(2023{\natexlab{b}})Li, Peng, Zhang, Ding, Hu, and
  Shen]{li2023deep}
W.~Li, Y.~Peng, M.~Zhang, L.~Ding, H.~Hu, and L.~Shen.
\newblock Deep model fusion: A survey.
\newblock \emph{arXiv preprint arXiv:2309.15698}, 2023{\natexlab{b}}.

\bibitem[Li et~al.(2019)Li, Huang, Yang, Wang, and Zhang]{li2019convergence}
X.~Li, K.~Huang, W.~Yang, S.~Wang, and Z.~Zhang.
\newblock On the convergence of fedavg on non-iid data.
\newblock In \emph{International Conference on Learning Representations}, 2019.

\bibitem[Li et~al.(2015)Li, Yosinski, Clune, Lipson, and
  Hopcroft]{li2015convergent}
Y.~Li, J.~Yosinski, J.~Clune, H.~Lipson, and J.~Hopcroft.
\newblock Convergent learning: Do different neural networks learn the same
  representations?
\newblock \emph{arXiv preprint arXiv:1511.07543}, 2015.

\bibitem[Liu et~al.(2022)Liu, Tam, Muqeeth, Mohta, Huang, Bansal, and
  Raffel]{liu2022tfew}
H.~Liu, D.~Tam, M.~Muqeeth, J.~Mohta, T.~Huang, M.~Bansal, and C.~A. Raffel.
\newblock Few-shot parameter-efficient fine-tuning is better and cheaper than
  in-context learning.
\newblock \emph{Advances in Neural Information Processing Systems},
  35:\penalty0 1950--1965, 2022.

\bibitem[Marneffe et~al.(2019)Marneffe, Simons, and Tonhauser]{cb}
M.-C.~d. Marneffe, M.~Simons, and J.~Tonhauser.
\newblock {The CommitmentBank}: Investigating projection in naturally occurring
  discourse.
\newblock \emph{Proceedings of Sinn und Bedeutung 23}, 2019.

\bibitem[Matena and Raffel(2021)]{matena2021merging}
M.~Matena and C.~Raffel.
\newblock Merging models with fisher-weighted averaging.
\newblock In \emph{Advances in Neural Information Processing Systems
  (NeurIPS)}, 2021.
\newblock \url{https://arxiv.org/abs/2111.09832}.

\bibitem[McMahan et~al.(2017)McMahan, Moore, Ramage, Hampson, and
  y~Arcas]{mcmahan2017communication}
B.~McMahan, E.~Moore, D.~Ramage, S.~Hampson, and B.~A. y~Arcas.
\newblock Communication-efficient learning of deep networks from decentralized
  data.
\newblock In \emph{Artificial intelligence and statistics}, pages 1273--1282.
  PMLR, 2017.

\bibitem[Netzer et~al.(2011)Netzer, Wang, Coates, Bissacco, Wu, and Ng]{svhn}
Y.~Netzer, T.~Wang, A.~Coates, A.~Bissacco, B.~Wu, and A.~Y. Ng.
\newblock Reading digits in natural images with unsupervised feature learning.
\newblock In \emph{Advances in Neural Information Processing Systems (NeurIPS)
  Workshops}, 2011.
\newblock
  \url{https://storage.googleapis.com/pub-tools-public-publication-data/pdf/37648.pdf}.

\bibitem[Neyshabur et~al.(2020)Neyshabur, Sedghi, and
  Zhang]{neyshabur2020being}
B.~Neyshabur, H.~Sedghi, and C.~Zhang.
\newblock What is being transferred in transfer learning?
\newblock \emph{Advances in neural information processing systems},
  33:\penalty0 512--523, 2020.

\bibitem[Nie et~al.(2019)Nie, Williams, Dinan, Bansal, Weston, and
  Kiela]{nie2019adversarial}
Y.~Nie, A.~Williams, E.~Dinan, M.~Bansal, J.~Weston, and D.~Kiela.
\newblock {Adversarial NLI}: A new benchmark for natural language
  understanding.
\newblock \emph{arXiv preprint arXiv:1910.14599}, 2019.

\bibitem[Orgad et~al.(2023)Orgad, Kawar, and Belinkov]{orgad2023editing}
H.~Orgad, B.~Kawar, and Y.~Belinkov.
\newblock Editing implicit assumptions in text-to-image diffusion models.
\newblock \emph{arXiv preprint arXiv:2303.08084}, 2023.

\bibitem[Ortiz{-}Jim{\'{e}}nez et~al.(2023)Ortiz{-}Jim{\'{e}}nez, Favero, and
  Frossard]{ortizjimenez2023tangent}
G.~Ortiz{-}Jim{\'{e}}nez, A.~Favero, and P.~Frossard.
\newblock Task arithmetic in the tangent space: Improved editing of pre-trained
  models.
\newblock \emph{NeurIPS}, 2023.
\newblock \url{https://arxiv.org/abs/2305:12827}.

\bibitem[Phang et~al.(2018)Phang, F{\'e}vry, and Bowman]{phang2018sentence}
J.~Phang, T.~F{\'e}vry, and S.~R. Bowman.
\newblock Sentence encoders on stilts: Supplementary training on intermediate
  labeled-data tasks.
\newblock \emph{arXiv preprint arXiv:1811.01088}, 2018.

\bibitem[Pilehvar and Camacho-Collados(2019)]{pilehvar2018wic}
M.~T. Pilehvar and J.~Camacho-Collados.
\newblock {WiC}: The word-in-context dataset for evaluating context-sensitive
  meaning representations.
\newblock In \emph{Proceedings of NAACL-HLT}, 2019.

\bibitem[Poth et~al.(2021)Poth, Pfeiffer, R{\"u}ckl{\'e}, and
  Gurevych]{poth-etal-2021-pre}
C.~Poth, J.~Pfeiffer, A.~R{\"u}ckl{\'e}, and I.~Gurevych.
\newblock {W}hat to pre-train on? {E}fficient intermediate task selection.
\newblock In \emph{Proceedings of the 2021 Conference on Empirical Methods in
  Natural Language Processing}, pages 10585--10605, Online and Punta Cana,
  Dominican Republic, Nov. 2021.

\bibitem[Pruksachatkun et~al.(2020)Pruksachatkun, Phang, Liu, Htut, Zhang,
  Pang, Vania, Kann, and Bowman]{pruksachatkun-etal-2020-intermediate}
Y.~Pruksachatkun, J.~Phang, H.~Liu, P.~M. Htut, X.~Zhang, R.~Y. Pang, C.~Vania,
  K.~Kann, and S.~R. Bowman.
\newblock Intermediate-task transfer learning with pretrained language models:
  When and why does it work?
\newblock In \emph{Proceedings of the 58th Annual Meeting of the Association
  for Computational Linguistics}, pages 5231--5247, Online, July 2020.

\bibitem[Radford et~al.(2021)Radford, Kim, Hallacy, Ramesh, Goh, Agarwal,
  Sastry, Askell, Mishkin, Clark, Krueger, and Sutskever]{radford2021learning}
A.~Radford, J.~W. Kim, C.~Hallacy, A.~Ramesh, G.~Goh, S.~Agarwal, G.~Sastry,
  A.~Askell, P.~Mishkin, J.~Clark, G.~Krueger, and I.~Sutskever.
\newblock Learning transferable visual models from natural language
  supervision.
\newblock In \emph{International Conference on Machine Learning (ICML)}, 2021.
\newblock \url{https://arxiv.org/abs/2103.00020}.

\bibitem[Raffel et~al.(2020{\natexlab{a}})Raffel, Shazeer, Roberts, Lee,
  Narang, Matena, Zhou, Li, and Liu]{colin2020exploring}
C.~Raffel, N.~Shazeer, A.~Roberts, K.~Lee, S.~Narang, M.~Matena, Y.~Zhou,
  W.~Li, and P.~J. Liu.
\newblock Exploring the limits of transfer learning with a unified text-to-text
  transformer.
\newblock \emph{Journal of Machine Learning Research (JMLR)},
  2020{\natexlab{a}}.
\newblock \url{http://jmlr.org/papers/v21/20-074.html}.

\bibitem[Raffel et~al.(2020{\natexlab{b}})Raffel, Shazeer, Roberts, Lee,
  Narang, Matena, Zhou, Li, and Liu]{raffel2019exploring}
C.~Raffel, N.~M. Shazeer, A.~Roberts, K.~Lee, S.~Narang, M.~Matena, Y.~Zhou,
  W.~Li, and P.~J. Liu.
\newblock Exploring the limits of transfer learning with a unified text-to-text
  transformer.
\newblock \emph{ArXiv}, abs/1910.10683, 2020{\natexlab{b}}.

\bibitem[Ram{\'e} et~al.(2022)Ram{\'e}, Ahuja, Zhang, Cord, Bottou, and
  Lopez-Paz]{rame2022modelrat}
A.~Ram{\'e}, K.~Ahuja, J.~Zhang, M.~Cord, L.~Bottou, and D.~Lopez-Paz.
\newblock Model ratatouille: Recycling diverse models for out-of-distribution
  generalization.
\newblock \emph{arXiv preprint arXiv:2212.10445}, 2022.

\bibitem[Ram{\'e} et~al.(2023)Ram{\'e}, Kirchmeyer, Rahier, Rakotomamonjy,
  Gallinari, and Cord]{Ram2022DiverseWA}
A.~Ram{\'e}, M.~Kirchmeyer, T.~Rahier, A.~Rakotomamonjy, P.~Gallinari, and
  M.~Cord.
\newblock Diverse weight averaging for out-of-distribution generalization.
\newblock \emph{ICML}, 2023.

\bibitem[Roemmele et~al.(2011)Roemmele, Bejan, and Gordon]{copa}
M.~Roemmele, C.~A. Bejan, and A.~S. Gordon.
\newblock Choice of plausible alternatives: An evaluation of commonsense causal
  reasoning.
\newblock \emph{2011 AAAI Spring Symposium Series}, 2011.

\bibitem[Rogers et~al.(2020)Rogers, Kovaleva, Downey, and
  Rumshisky]{quail_dataset}
A.~Rogers, O.~Kovaleva, M.~Downey, and A.~Rumshisky.
\newblock Getting closer to {AI} complete question answering: {A} set of
  prerequisite real tasks.
\newblock In \emph{The Thirty-Fourth {AAAI} Conference on Artificial
  Intelligence, {AAAI} 2020, The Thirty-Second Innovative Applications of
  Artificial Intelligence Conference, {IAAI} 2020, The Tenth {AAAI} Symposium
  on Educational Advances in Artificial Intelligence, {EAAI} 2020, New York,
  NY, USA, February 7-12, 2020}, pages 8722--8731. {AAAI} Press, 2020.
\newblock URL \url{https://aaai.org/ojs/index.php/AAAI/article/view/6398}.

\bibitem[Ruder(2016)]{ruder2016overview}
S.~Ruder.
\newblock An overview of gradient descent optimization algorithms.
\newblock \emph{arXiv preprint arXiv:1609.04747}, 2016.

\bibitem[Sakaguchi et~al.(2020)Sakaguchi, Le~Bras, Bhagavatula, and
  Choi]{sakaguchi2020winogrande}
K.~Sakaguchi, R.~Le~Bras, C.~Bhagavatula, and Y.~Choi.
\newblock Winogrande: An adversarial winograd schema challenge at scale.
\newblock In \emph{Proceedings of the AAAI Conference on Artificial
  Intelligence}, 2020.

\bibitem[Sanh et~al.(2021{\natexlab{a}})Sanh, Webson, Raffel, Bach, Sutawika,
  Alyafeai, Chaffin, Stiegler, Scao, Raja, et~al.]{sanh2021multitask}
V.~Sanh, A.~Webson, C.~Raffel, S.~H. Bach, L.~Sutawika, Z.~Alyafeai,
  A.~Chaffin, A.~Stiegler, T.~L. Scao, A.~Raja, et~al.
\newblock Multitask prompted training enables zero-shot task generalization.
\newblock \emph{arXiv preprint arXiv:2110.08207}, 2021{\natexlab{a}}.

\bibitem[Sanh et~al.(2021{\natexlab{b}})Sanh, Webson, Raffel, Bach, Sutawika,
  Alyafeai, Chaffin, Stiegler, Scao, Raja, et~al.]{sanh2022multitask}
V.~Sanh, A.~Webson, C.~Raffel, S.~H. Bach, L.~Sutawika, Z.~Alyafeai,
  A.~Chaffin, A.~Stiegler, T.~L. Scao, A.~Raja, et~al.
\newblock Multitask prompted training enables zero-shot task generalization.
\newblock In \emph{International Conference on Learning Representations
  (ICLR)}, 2021{\natexlab{b}}.
\newblock \url{https://arxiv.org/abs/2110.08207}.

\bibitem[Sap et~al.(2019)Sap, Rashkin, Chen, Le~Bras, and Choi]{sap2019social}
M.~Sap, H.~Rashkin, D.~Chen, R.~Le~Bras, and Y.~Choi.
\newblock Social iqa: Commonsense reasoning about social interactions.
\newblock In \emph{Proceedings of the 2019 Conference on Empirical Methods in
  Natural Language Processing and the 9th International Joint Conference on
  Natural Language Processing (EMNLP-IJCNLP)}, pages 4463--4473, 2019.

\bibitem[Sharma et~al.(2018)Sharma, Allen, Bakhshandeh, and
  Mostafazadeh]{sharma2018tackling}
R.~Sharma, J.~Allen, O.~Bakhshandeh, and N.~Mostafazadeh.
\newblock Tackling the story ending biases in the story cloze test.
\newblock In \emph{Proceedings of the 56th Annual Meeting of the Association
  for Computational Linguistics (Volume 2: Short Papers)}, pages 752--757,
  2018.

\bibitem[Shnarch et~al.(2022)Shnarch, Halfon, Gera, Danilevsky, Katsis,
  Choshen, Cooper, Epelboim, Zhang, Wang, et~al.]{shnarch2022label}
E.~Shnarch, A.~Halfon, A.~Gera, M.~Danilevsky, Y.~Katsis, L.~Choshen, M.~S.
  Cooper, D.~Epelboim, Z.~Zhang, D.~Wang, et~al.
\newblock Label sleuth: From unlabeled text to a classifier in a few hours.
\newblock In \emph{Conference on Empirical Methods in Natural Language
  Processing}, 2022.

\bibitem[Singh and Jaggi(2020)]{singh2020model}
S.~P. Singh and M.~Jaggi.
\newblock Model fusion via optimal transport.
\newblock \emph{Advances in Neural Information Processing Systems},
  33:\penalty0 22045--22055, 2020.

\bibitem[Stallkamp et~al.(2011)Stallkamp, Schlipsing, Salmen, and Igel]{gtsrb}
J.~Stallkamp, M.~Schlipsing, J.~Salmen, and C.~Igel.
\newblock The german traffic sign recognition benchmark: a multi-class
  classification competition.
\newblock In \emph{International Joint Conference on Neural Networks (IJCNN)},
  2011.
\newblock \url{https://ieeexplore.ieee.org/document/6033395}.

\bibitem[Sung et~al.(2023)Sung, Li, Lin, Gan, Bansal, and
  Wang]{Sung2023AnEmpiricalSO}
Y.-L. Sung, L.~Li, K.~Lin, Z.~Gan, M.~Bansal, and L.~Wang.
\newblock An empirical study of multimodal model merging.
\newblock \emph{Empirical Methods in Natural Language Processing (Findings)},
  2023.

\bibitem[Tafjord et~al.(2019)Tafjord, Gardner, Lin, and
  Clark]{tafjord2019quartz}
O.~Tafjord, M.~Gardner, K.~Lin, and P.~Clark.
\newblock Quartz: An open-domain dataset of qualitative relationship questions.
\newblock In \emph{Proceedings of the 2019 Conference on Empirical Methods in
  Natural Language Processing and the 9th International Joint Conference on
  Natural Language Processing (EMNLP-IJCNLP)}, pages 5941--5946, 2019.

\bibitem[Tatro et~al.(2020)Tatro, Chen, Das, Melnyk, Sattigeri, and
  Lai]{tatro2020optimizing}
N.~Tatro, P.-Y. Chen, P.~Das, I.~Melnyk, P.~Sattigeri, and R.~Lai.
\newblock Optimizing mode connectivity via neuron alignment.
\newblock \emph{Advances in Neural Information Processing Systems},
  33:\penalty0 15300--15311, 2020.

\bibitem[Tay et~al.(2022)Tay, Dehghani, Tran, Garcia, Wei, Wang, Chung, Bahri,
  Schuster, Zheng, et~al.]{tay2022ul2}
Y.~Tay, M.~Dehghani, V.~Q. Tran, X.~Garcia, J.~Wei, X.~Wang, H.~W. Chung,
  D.~Bahri, T.~Schuster, S.~Zheng, et~al.
\newblock Ul2: Unifying language learning paradigms.
\newblock In \emph{The Eleventh International Conference on Learning
  Representations}, 2022.

\bibitem[Thimm and Fiesler(1995)]{thimm1995evaluatingtopk}
G.~Thimm and E.~Fiesler.
\newblock Evaluating pruning methods.
\newblock In \emph{International Symposium on Artificial Neural Networks},
  1995.

\bibitem[Vaswani et~al.(2017)Vaswani, Shazeer, Parmar, Uszkoreit, Jones, Gomez,
  Kaiser, and Polosukhin]{vaswani2017attention}
A.~Vaswani, N.~Shazeer, N.~Parmar, J.~Uszkoreit, L.~Jones, A.~N. Gomez,
  {\L}.~Kaiser, and I.~Polosukhin.
\newblock Attention is all you need.
\newblock \emph{Advances in Neural Information Processing Systems (NeurIPS)},
  2017.
\newblock \url{https://arxiv.org/abs/1706.03762}.

\bibitem[Wang et~al.(2018)Wang, Singh, Michael, Hill, Levy, and
  Bowman]{wang2018glue}
A.~Wang, A.~Singh, J.~Michael, F.~Hill, O.~Levy, and S.~R. Bowman.
\newblock Glue: A multi-task benchmark and analysis platform for natural
  language understanding.
\newblock In \emph{International Conference on Learning Representations
  (ICLR)}, 2018.
\newblock \url{https://arxiv.org/abs/1804.07461}.

\bibitem[Wang et~al.(2020)Wang, Yurochkin, Sun, Papailiopoulos, and
  Khazaeni]{wang2020federated}
H.~Wang, M.~Yurochkin, Y.~Sun, D.~Papailiopoulos, and Y.~Khazaeni.
\newblock Federated learning with matched averaging.
\newblock In \emph{International Conference on Learning Representations}, 2020.

\bibitem[Weller et~al.(2022)Weller, Seppi, and Gardner]{weller-etal-2022-use}
O.~Weller, K.~Seppi, and M.~Gardner.
\newblock When to use multi-task learning vs intermediate fine-tuning for
  pre-trained encoder transfer learning.
\newblock In \emph{Proceedings of the 60th Annual Meeting of the Association
  for Computational Linguistics (Volume 2: Short Papers)}, pages 272--282,
  Dublin, Ireland, May 2022.

\bibitem[Wolf et~al.(2019)Wolf, Debut, Sanh, Chaumond, Delangue, Moi, Cistac,
  Rault, Louf, Funtowicz, et~al.]{wolf2019huggingface}
T.~Wolf, L.~Debut, V.~Sanh, J.~Chaumond, C.~Delangue, A.~Moi, P.~Cistac,
  T.~Rault, R.~Louf, M.~Funtowicz, et~al.
\newblock Huggingface's transformers: State-of-the-art natural language
  processing, 2019.
\newblock \url{https://arxiv.org/abs/1910.03771}.

\bibitem[Wortsman et~al.(2022{\natexlab{a}})Wortsman, Ilharco, Gadre, Roelofs,
  Gontijo-Lopes, Morcos, Namkoong, Farhadi, Carmon, Kornblith,
  et~al.]{wortsman2022model}
M.~Wortsman, G.~Ilharco, S.~Y. Gadre, R.~Roelofs, R.~Gontijo-Lopes, A.~S.
  Morcos, H.~Namkoong, A.~Farhadi, Y.~Carmon, S.~Kornblith, et~al.
\newblock Model soups: averaging weights of multiple fine-tuned models improves
  accuracy without increasing inference time.
\newblock In \emph{International Conference on Machine Learning (ICML)},
  2022{\natexlab{a}}.
\newblock \url{https://arxiv.org/abs/2203.05482}.

\bibitem[Wortsman et~al.(2022{\natexlab{b}})Wortsman, Ilharco, Li, Kim,
  Hajishirzi, Farhadi, Namkoong, and Schmidt]{wortsman2021robust}
M.~Wortsman, G.~Ilharco, M.~Li, J.~W. Kim, H.~Hajishirzi, A.~Farhadi,
  H.~Namkoong, and L.~Schmidt.
\newblock Robust fine-tuning of zero-shot models.
\newblock In \emph{Conference on Computer Vision and Pattern Recognition
  (CVPR)}, 2022{\natexlab{b}}.
\newblock \url{https://arxiv.org/abs/2109.01903}.

\bibitem[Xiao et~al.(2016)Xiao, Ehinger, Hays, Torralba, and Oliva]{sun397}
J.~Xiao, K.~A. Ehinger, J.~Hays, A.~Torralba, and A.~Oliva.
\newblock Sun database: Exploring a large collection of scene categories.
\newblock \emph{International Journal of Computer Vision (IJCV)}, 2016.
\newblock \url{https://link.springer.com/article/10.1007/s11263-014-0748-y}.

\bibitem[Yadav and Bansal(2023)]{yadav2023exssnet}
P.~Yadav and M.~Bansal.
\newblock Exclusive supermask subnetwork training for continual learning.
\newblock In \emph{Findings of the Association for Computational Linguistics:
  ACL 2023}, pages 569--587, Toronto, Canada, July 2023. Association for
  Computational Linguistics.
\newblock \doi{10.18653/v1/2023.findings-acl.36}.
\newblock URL \url{https://aclanthology.org/2023.findings-acl.36}.

\bibitem[Yadav et~al.(2023)Yadav, Sun, Ding, Li, Zhang, Tan, Bhatia, Ma,
  Nallapati, Ramanathan, Bansal, and Xiang]{yadav2023codecl}
P.~Yadav, Q.~Sun, H.~Ding, X.~Li, D.~Zhang, M.~Tan, P.~Bhatia, X.~Ma,
  R.~Nallapati, M.~K. Ramanathan, M.~Bansal, and B.~Xiang.
\newblock Exploring continual learning for code generation models.
\newblock In \emph{Proceedings of the 61st Annual Meeting of the Association
  for Computational Linguistics (Volume 2: Short Papers)}, pages 782--792,
  Toronto, Canada, July 2023. Association for Computational Linguistics.
\newblock \doi{10.18653/v1/2023.acl-short.68}.
\newblock URL \url{https://aclanthology.org/2023.acl-short.68}.

\bibitem[Yang et~al.(2015)Yang, Yih, and Meek]{yang-etal-2015-wikiqa}
Y.~Yang, W.-t. Yih, and C.~Meek.
\newblock {W}iki{QA}: A challenge dataset for open-domain question answering.
\newblock In \emph{Proceedings of the 2015 Conference on Empirical Methods in
  Natural Language Processing}, pages 2013--2018, Lisbon, Portugal, Sept. 2015.
  Association for Computational Linguistics.
\newblock \doi{10.18653/v1/D15-1237}.
\newblock URL \url{https://aclanthology.org/D15-1237}.

\bibitem[Zellers et~al.(2019)Zellers, Holtzman, Bisk, Farhadi, and
  Choi]{zellers2019hellaswag}
R.~Zellers, A.~Holtzman, Y.~Bisk, A.~Farhadi, and Y.~Choi.
\newblock {HellaSwag}: Can a machine really finish your sentence?
\newblock \emph{arXiv preprint arXiv:1905.07830}, 2019.

\bibitem[Zhang and Yang(2021)]{zhang2021survey}
Y.~Zhang and Q.~Yang.
\newblock A survey on multi-task learning.
\newblock \emph{IEEE Transactions on Knowledge and Data Engineering},
  34\penalty0 (12):\penalty0 5586--5609, 2021.

\bibitem[Zhang et~al.(2019)Zhang, Baldridge, and He]{paws2019naacl}
Y.~Zhang, J.~Baldridge, and L.~He.
\newblock {PAWS: Paraphrase Adversaries from Word Scrambling}.
\newblock In \emph{Proc. of NAACL}, 2019.

\bibitem[Zhuang et~al.(2020)Zhuang, Qi, Duan, Xi, Zhu, Zhu, Xiong, and
  He]{zhuang2020comprehensive}
F.~Zhuang, Z.~Qi, K.~Duan, D.~Xi, Y.~Zhu, H.~Zhu, H.~Xiong, and Q.~He.
\newblock A comprehensive survey on transfer learning.
\newblock \emph{Proceedings of the IEEE}, 2020.
\newblock \url{https://arxiv.org/abs/1911.02685}.

\end{thebibliography}

\newpage
\appendix
\vspace{20pt}

\textbf{{\Large Appendix for \methodshort{}}}

\section{Limitations and Future Works}
\label{sec:limitation}
Our works share the same general limitations of existing merging methods, like (1) a limited theoretical understanding of why and when weight interpolation works, what are the important underlying factors, and its proper connections with mode connectivity. Recent works like \cite{orgad2023editing} have demonstrated interesting relationships between weight disentanglement and mergingability of models; (2) that merging relies on common initialization and model architecture; and (3) merging individual-task models to create a multitask still lags behind the simultaneous multitask training. Moreover, it is not clear how to select the checkpoints for merging in order to create multitask models useful for specific domains. In addition, while our method provides a way to choose signs when merging task vectors, we still find that using the signs from a multitask model performs better. Some potential future works include figuring out a good way to estimate multitask signs without having access to the multitask model as this has the potential to bridge the gap between multitask merging and multitask training (as demonstrated in Section~\ref{analysis:estimating_sign}).

\section{Additional Results}
\label{sec:app_main}

\begin{table}[tbh!]
        \centering
        \begin{tabular}{lcccl}
        \toprule
        \textbf{Method} & \multicolumn{3}{c}{\textbf{Estimating Sign}} & \multicolumn{1}{l}{\textbf{Average}} \\
        \cmidrule(lr){2-4}
        & \textbf{Multitask} & \textbf{Samples} & \textbf{Init.} & \\
        \midrule
        \rowcolor{gray!20} \textbf{Fine-Tuned} & - & - & - & 71.4 \\
        \rowcolor{gray!20} \textbf{Multitask} & - & - & - & 73.1  \\
        \midrule
        \rowcolor{gray!20} \textbf{Averaging}~\cite{choshen2022fusing,wortsman2022model} & - & - & - & 58.0 \\
        \rowcolor{gray!20} \textbf{Task Vectors}~\cite{ilharco2023editing} & - & - & - & 63.9 \\
        \rowcolor{gray!20} \textbf{\methodshort{}} & - & - & - & \textbf{66.4} \\
        
        \midrule
        \multirow{3}{*}{\textbf{\methodshort{}}} & \greencmark & 32 & scratch & 66.5 [\textcolor{color2}{+0.1}] \\
        & \greencmark & 32 & mean & 67.7 [\textcolor{color2}{+1.2}]  \\
        & \greencmark & All & scratch & \textbf{72.0} [\textcolor{color2}{+5.6}] \\

        \bottomrule
        \end{tabular}
        \vspace{10pt}
        \captionsetup{type=table}
        \caption{\label{tab:app_oracle} \textbf{Merging Performance can be improved by estimating the Sign Vector by performing few-shot multitask training.} 
        We use the estimated sign as the elected sign and perform merging. 
        }
\end{table}

\subsection{Enhancing Performance by Estimating the Multitask Sign Vector.}
\label{sec:app_estimating_multitask_sign}
Considering the findings, we inquire whether it is possible to efficiently acquire multitask sign vectors without extensive multitask training. Our proposal involves utilizing a limited number of validation samples from each task to cheaply train a multitask model and subsequently derive the relevant sign vector. We create two multitask (IA)$^3$ models: one developed from scratch and another initialized using the average of task-specific (IA)$^3$ models intended for merging. We use 32 validation examples from each task to train this model.
In Table \ref{tab:oracle}, we notice using the sign vector from the fewshot multitask model initialized with mean yielded a performance increase of $3.8\%$ and $1.3\%$ compared to Task Arithmetic and \methodshort{}. Interestingly, training fewshot multitask training from scratch did not yield significant improvements over \methodshort{} without sign estimation. We believe that exploring this area further may result in improved merging techniques.

\begin{figure}[t!]
  \centering
  \begin{minipage}[t]{0.32\linewidth}
    \centering
    \includegraphics[width=1.2\linewidth]{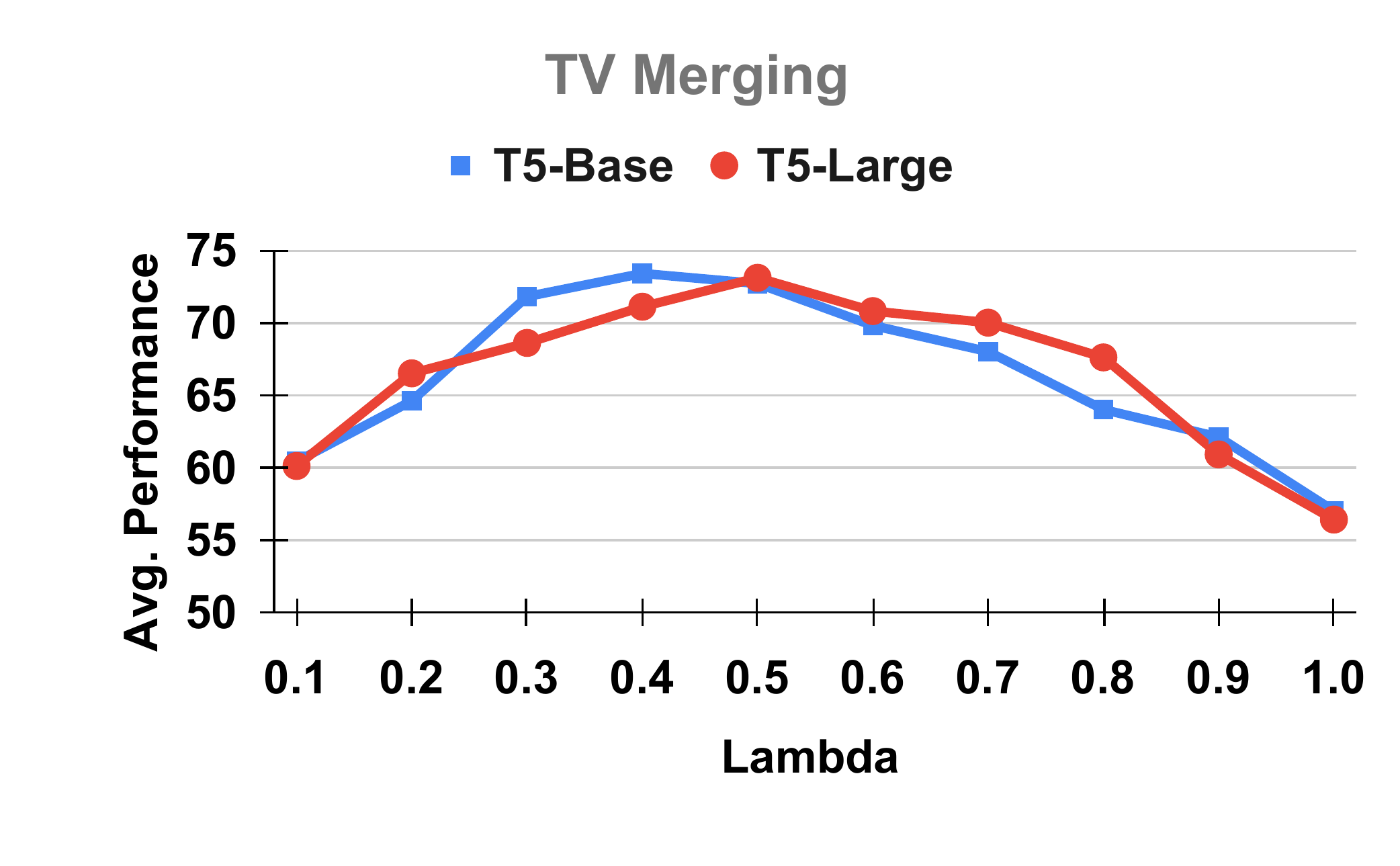}
  \end{minipage}
  \hfill
  \begin{minipage}[t]{0.32\linewidth}
    \centering
    \includegraphics[width=1.2\linewidth]{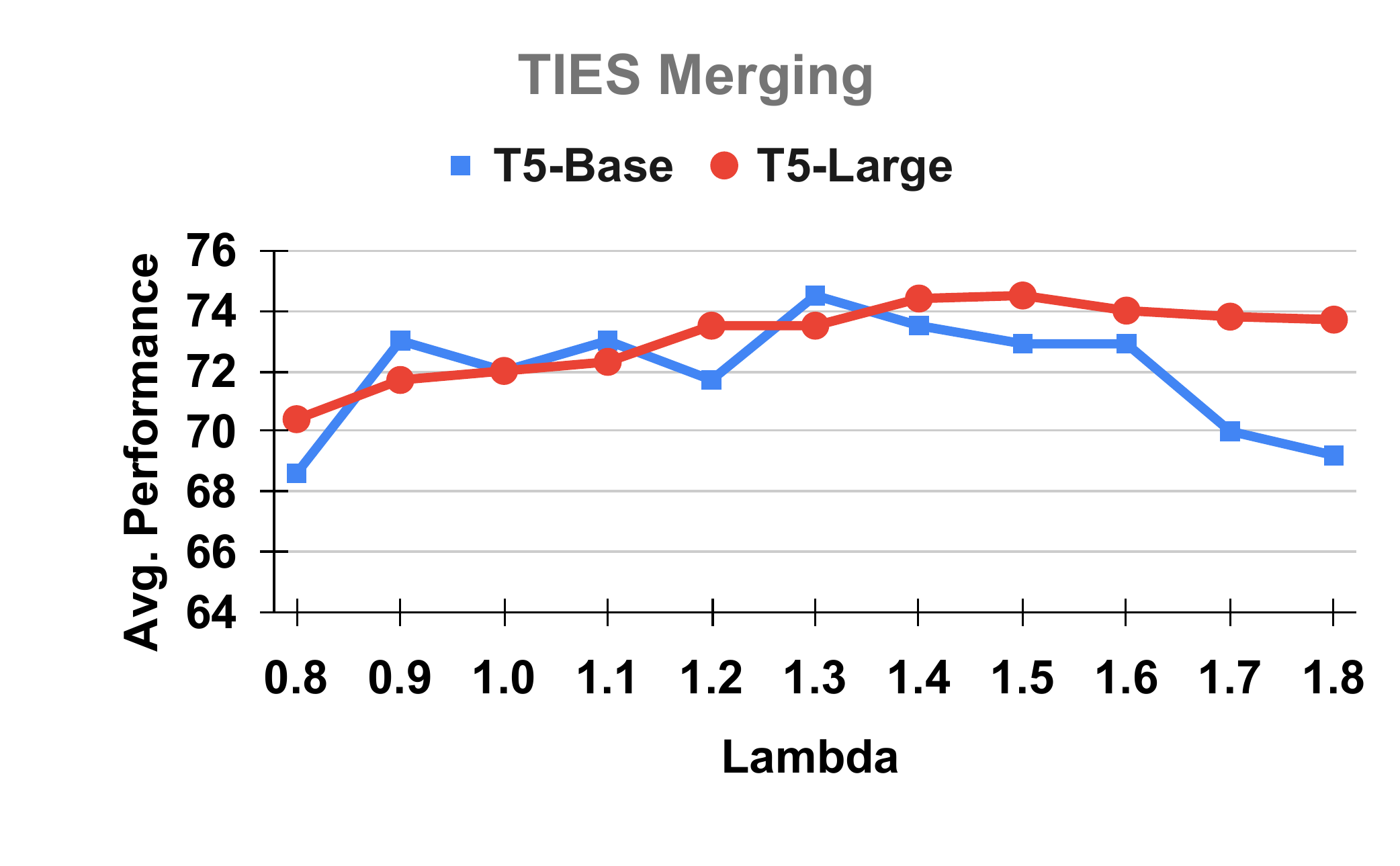}
  \end{minipage}
  \hfill
  \begin{minipage}[t]{0.32\linewidth}
    \centering
    \includegraphics[width=1\linewidth]{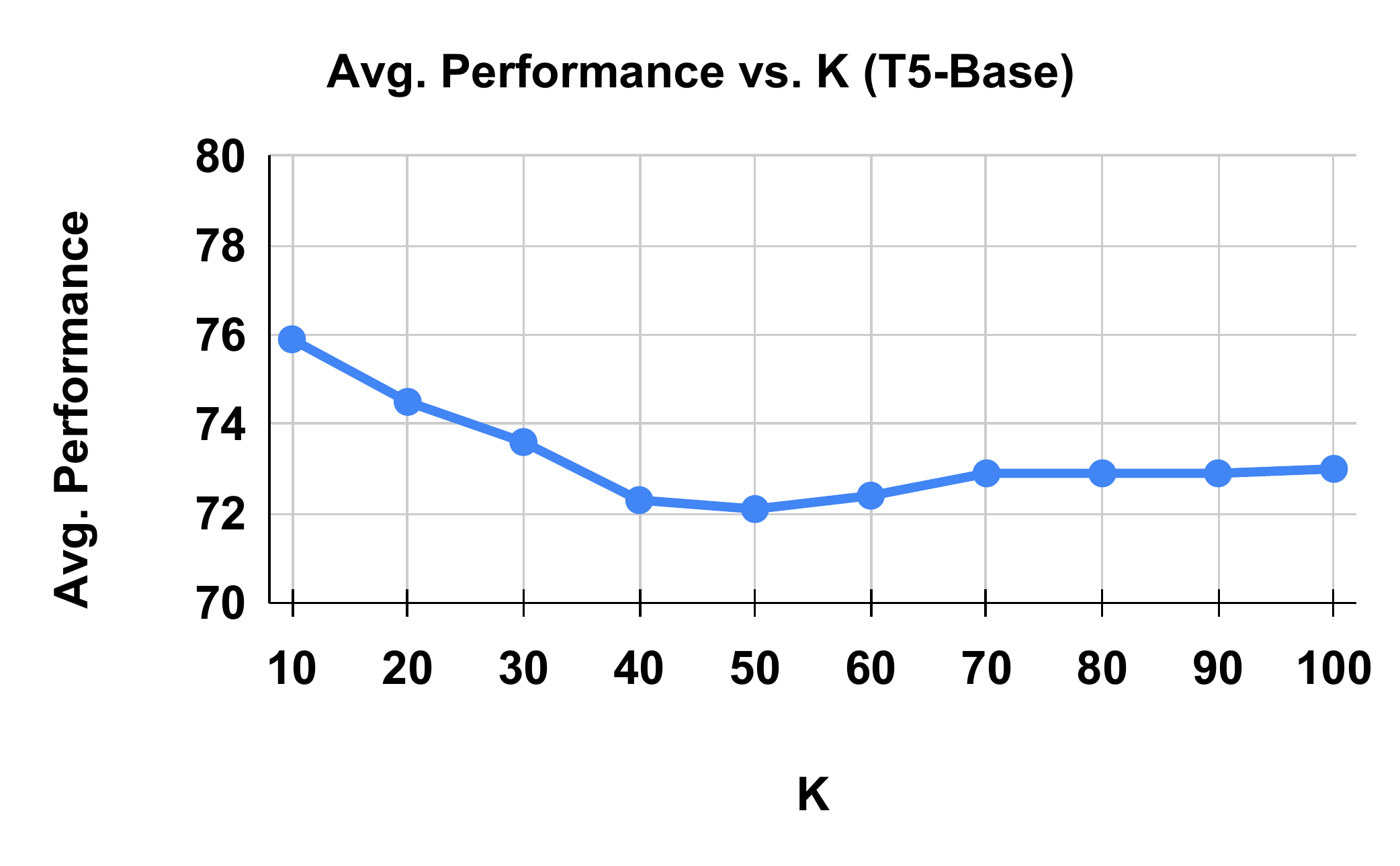}
  \end{minipage}
  \caption{\label{fig:hyperparams}\textbf{Performance as a function of hyperparameters}. For more details please refer to the response to our \textbf{general response.}}
\end{figure}

\subsection{Effect of Hyper-Parameters $\lambda$ and K on the Performance.} 
In Figure~\ref{fig:hyperparams} (left and middle), we plot the effect of $\lambda$ on the performance when merging T5-base and T5-large models trained on GLUE (Similar to Table-\ref{tab:main}). For \methodshort{}, we vary around the value 1 because TIES takes the mean of task vectors, whereas task arithmetic adds up the task vectors. Hence, a value of 1 for TIES is similar to using $\frac{1}{\# tasks}$ for Task Arithmetic~\cite{ilharco2023editing}. The range of 0.8-1.8 for TIES was selected based on preliminary experiments on the PEFT setting (as mentioned in Section~\ref{sec:exp_setup}). We find that \methodshort{} is much less sensitive to changes in (with an accuracy range of 68-75\% across the considered values of $\lambda$) compared to Task Arithmetic (with an accuracy range of 55-75). Figure~\ref{fig:hyperparams} (right) demonstrates the effect of $k$ as we increment the value of $k$ in steps of $10$ and skip $k=0$ as that would select no parameters. We observe that as $k$ increases the performance drops and then saturates. However, we would like to note that this curve might change based on the distribution of the parameter values in the task vector.

\begin{figure}[t!]
  \centering
  \begin{minipage}[t]{0.32\linewidth}
    \centering
    \includegraphics[width=\linewidth]{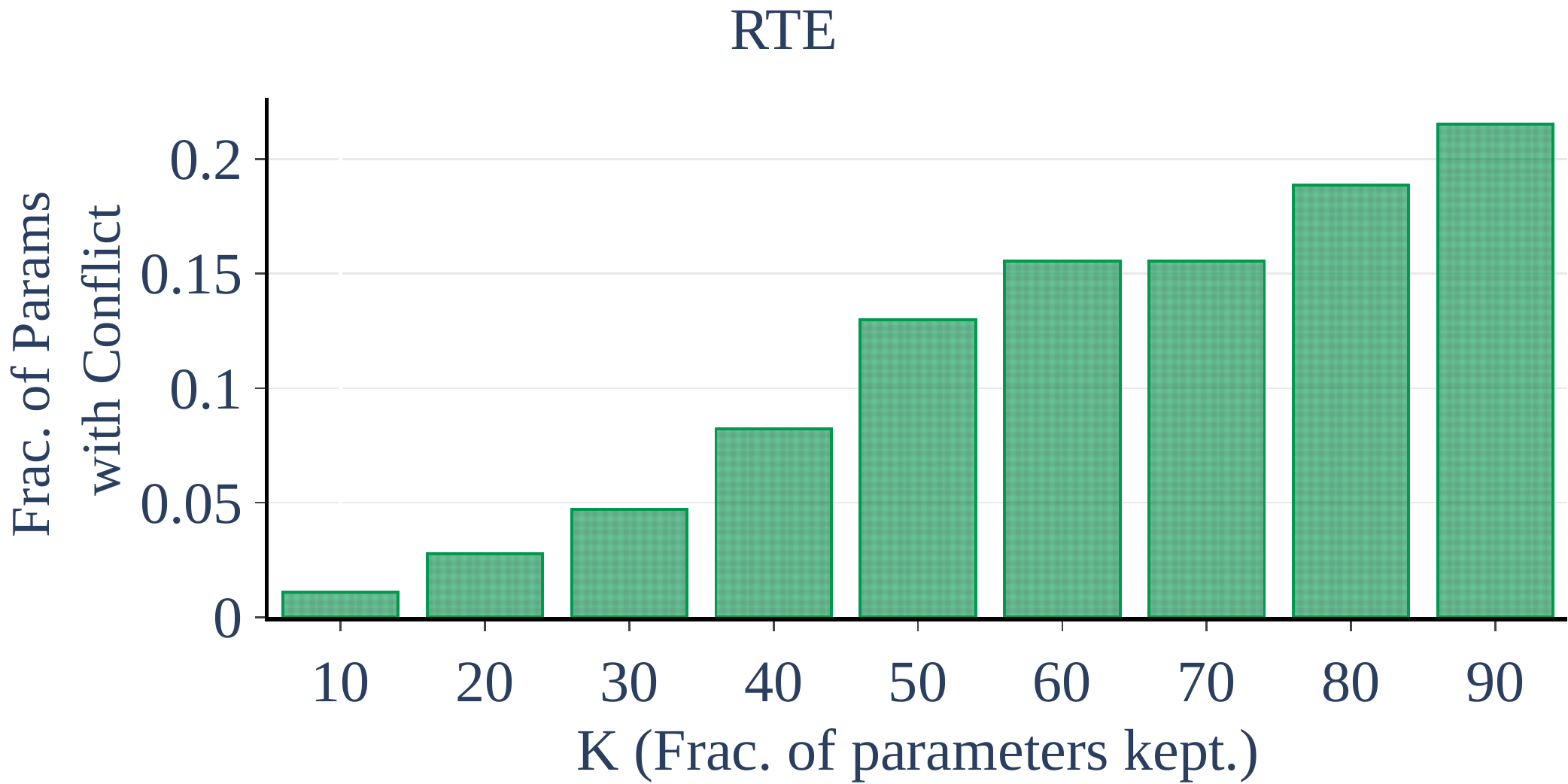}
  \end{minipage}
  \hfill
  \begin{minipage}[t]{0.32\linewidth}
    \centering
    \includegraphics[width=\linewidth]{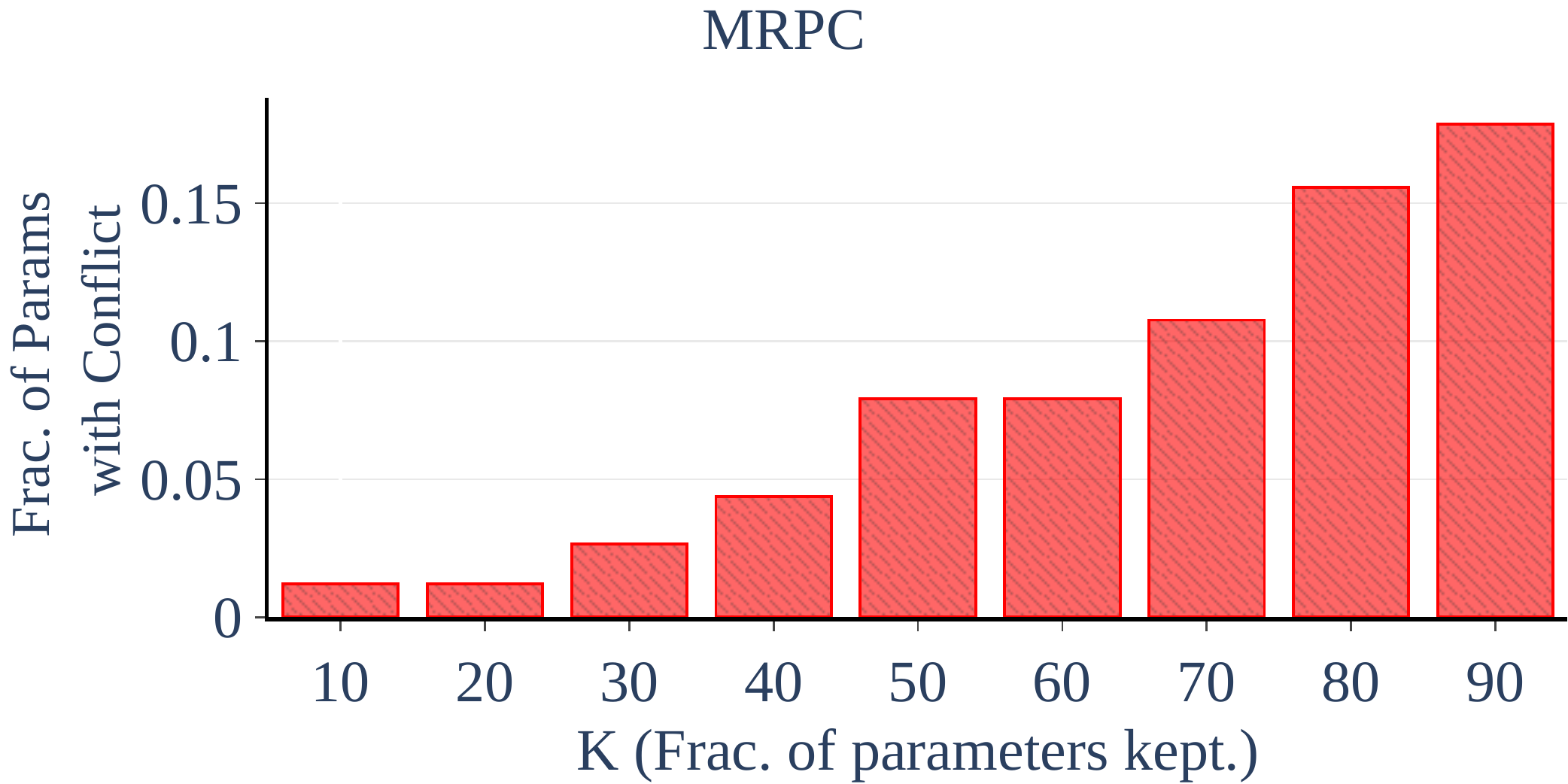}
  \end{minipage}
  \hfill
  \begin{minipage}[t]{0.32\linewidth}
    \centering
    \includegraphics[width=\linewidth]{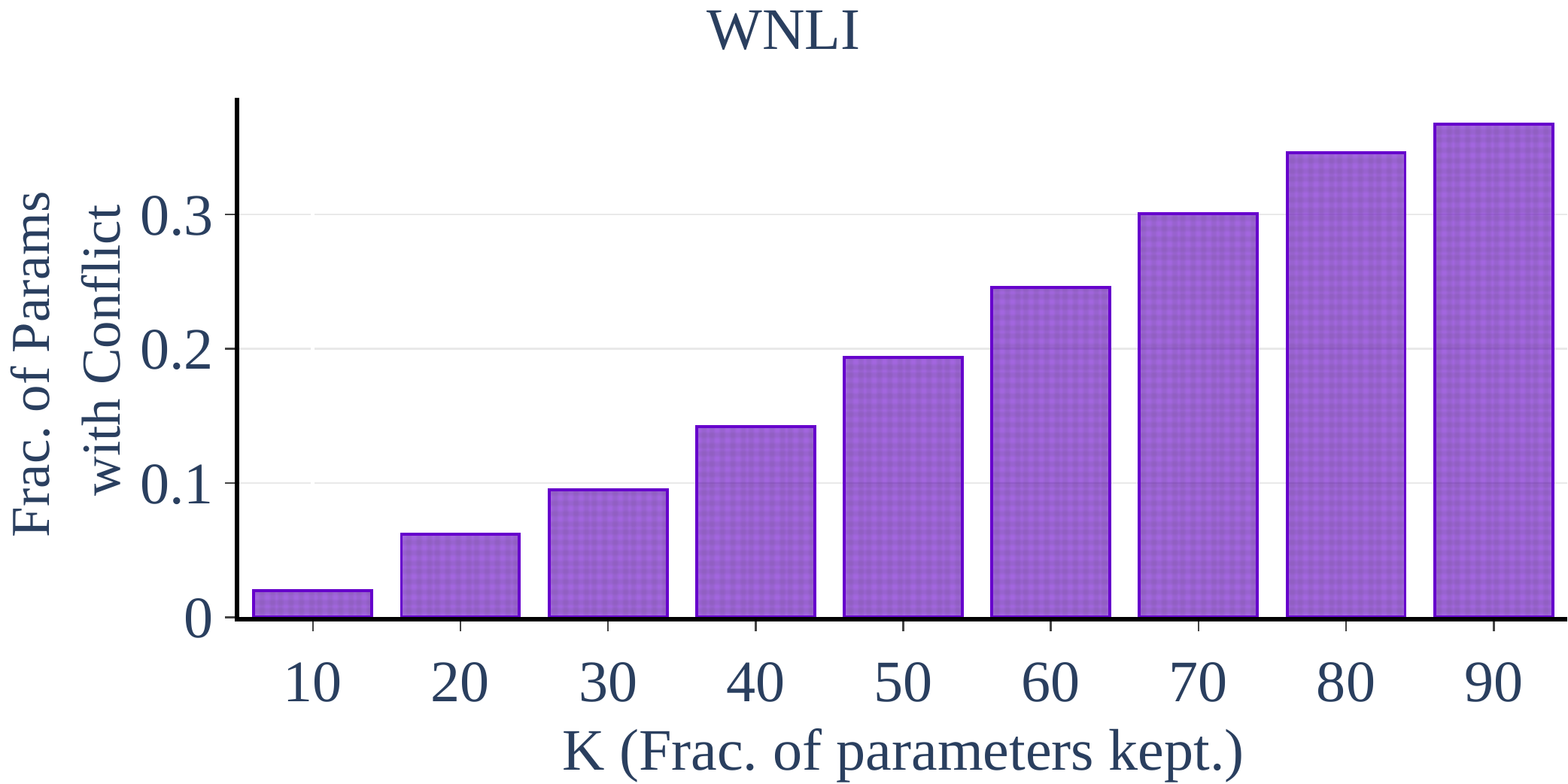}
  \end{minipage}
  \caption{\label{fig:conflict_vs_k}\textbf{Sign conflict increases as we trim less parameters.} For each task, we merge 10 different checkpoints from hunggingface hub and plot the sign conflict as a function of keeping only the top-k\% parameters.}
\end{figure}

\subsection{Sign Conflict Increases as We Trim Less Parameters}
In Figure~\ref{fig:conflict_vs_k}, we merge 10 \texttt{bert-base-uncased} checkpoints from huggingface finetuned on for three different glue tasks (RTE, MRPC, and WNLI) and plot the sign conflict as a function of $k$. As we keep more and more parameters, the sign conflict increases and reaches almost 80\%. This is also expected as there are many more nonzero parameters that can create conflict even if their magnitude is small.

\begin{figure}[t!]
  \centering
    \begin{minipage}[t]{0.24\linewidth}
    \centering
    \includegraphics[width=\linewidth]{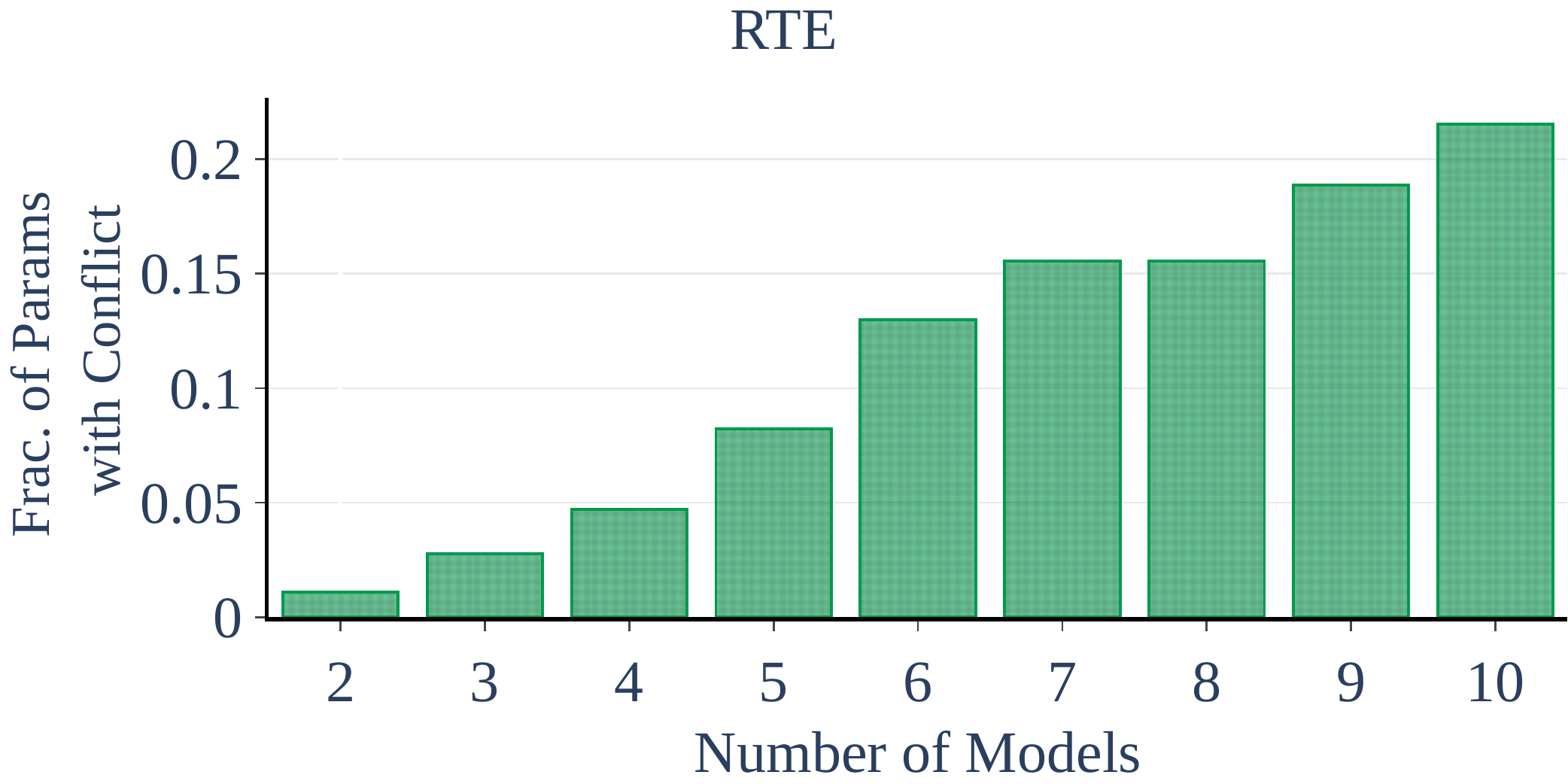}
  \end{minipage}
  \hfill
  \begin{minipage}[t]{0.24\linewidth}
    \centering
    \includegraphics[width=\linewidth]{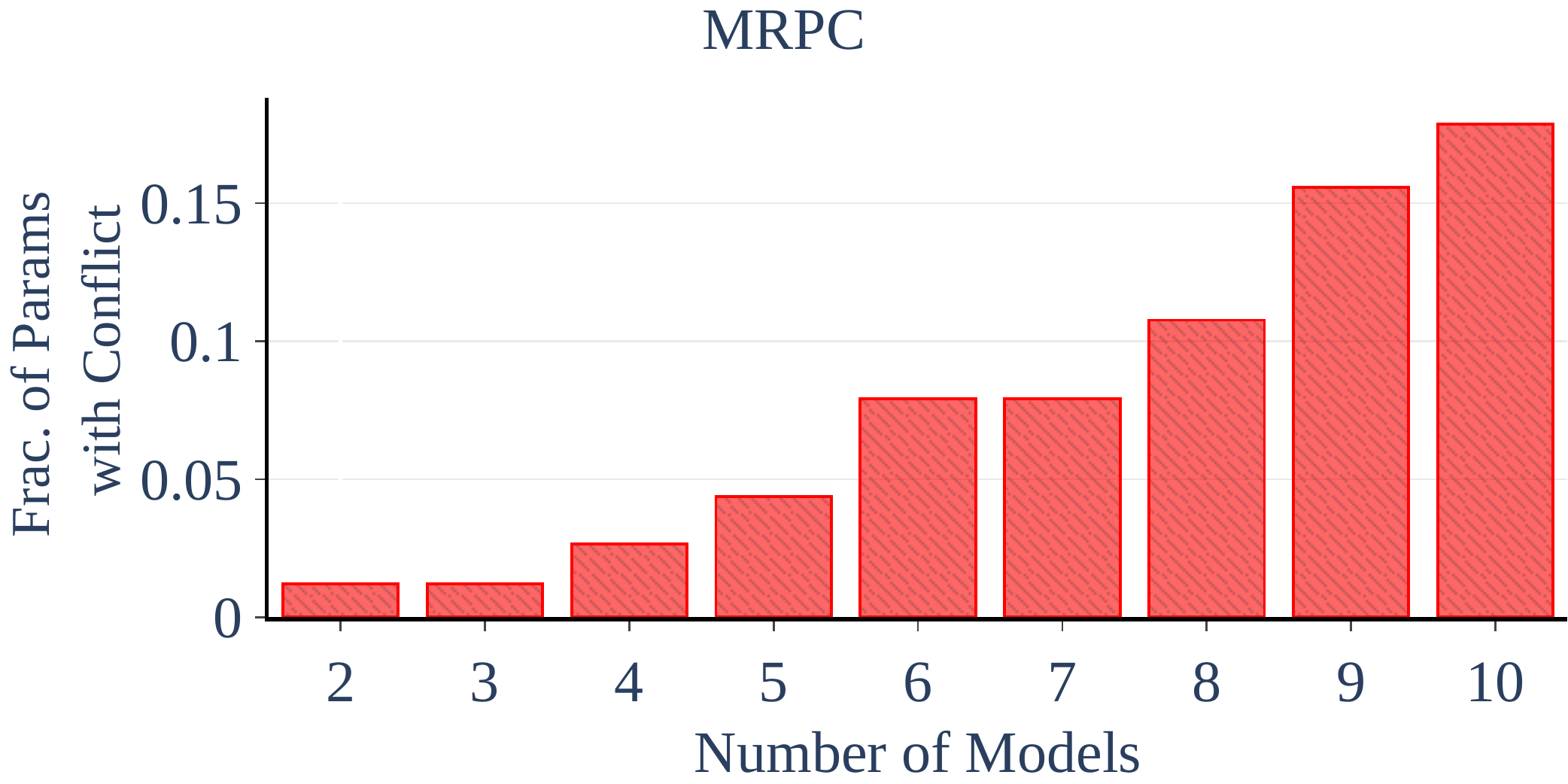}
  \end{minipage}
  \hfill
  \begin{minipage}[t]{0.24\linewidth}
    \centering
    \includegraphics[width=\linewidth]{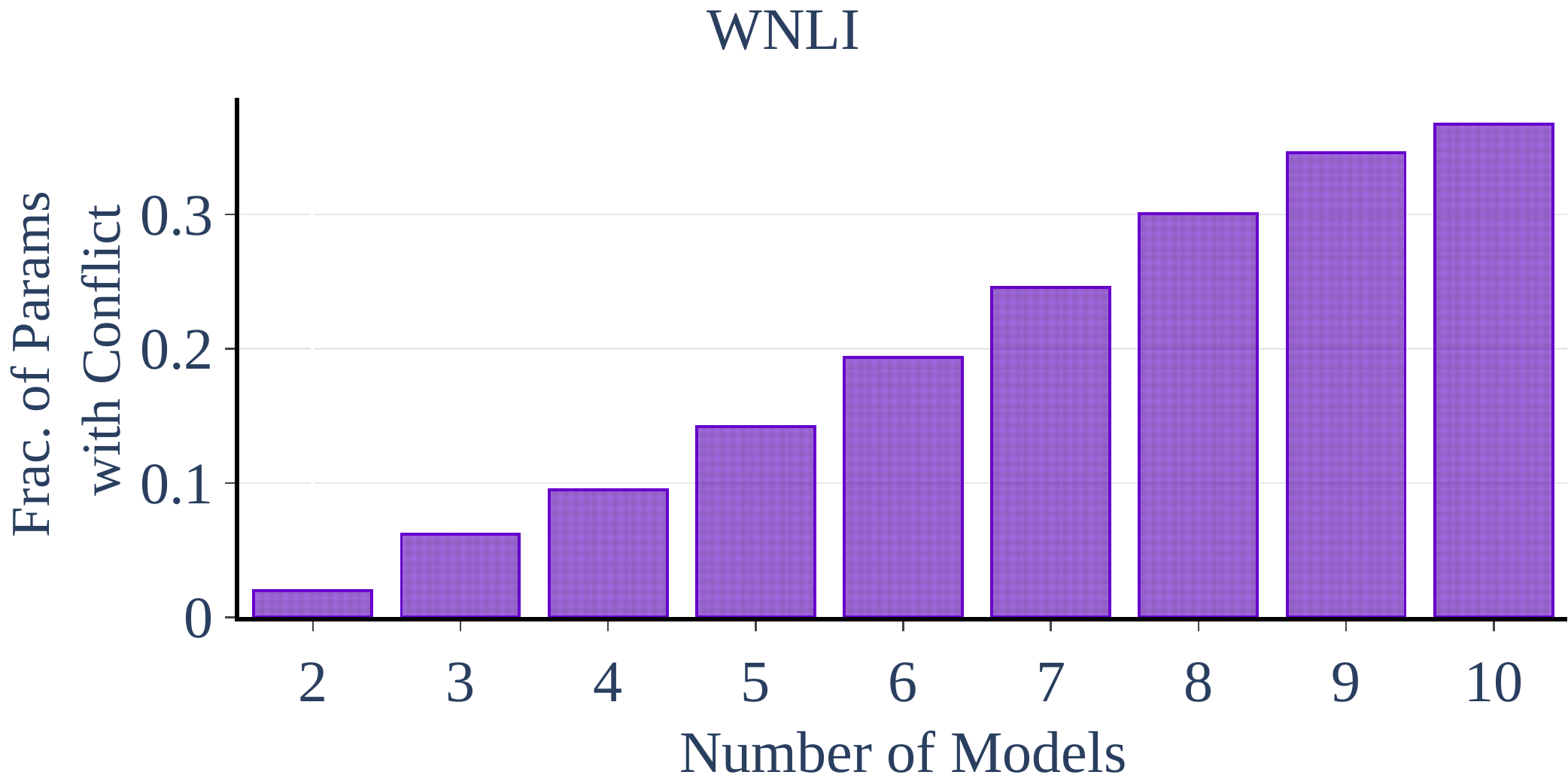}
  \end{minipage}
    \hfill
  \begin{minipage}[t]{0.24\linewidth}
    \centering
    \includegraphics[width=\linewidth]{figures/ia3_conflict_scale.pdf}
  \end{minipage}
\caption{\label{fig:same_task_interference}\textbf{Sign Conflict exists even when merging multiple checkpoints for the same task.} The first three plots are for RTE, MRPC, WNLI datasets when merging 10 Huggingface checkpoints, while the last one is when merging different tasks (Figure~\ref{fig:imp_conflict} from the main paper).}
\end{figure}

\subsection{Sign Conflicts Exists Between Different Checkpoints for the Same Task}
\label{sec:app_same_task_interference}
In Figure~\ref{fig:same_task_interference}, we show that sign conflicts exist even within models trained on the same task. We plotted the sign conflict (similar to Figure~\ref{fig:imp_conflict}) between the 10 checkpoints of RTE, MRPC, and WNLI from Huggingface. As the number of checkpoints increases, sign conflict increases. We also compare this with the sign interference when merging different task checkpoints and find a similar degree of interference in all of these cases. Hence, sign conflicts exist even within models trained on the same dataset. We suspect that this is because models are highly overparameterized and hence there are multiple subnetworks (subsets of parameters) that could lead to the same performance where different finetuning runs update the same parameters in different directions.

\begin{figure}[tbh!]
  \centering
  \begin{subfigure}[b]{0.48\linewidth}
        \centering
        \includegraphics[width=\linewidth]{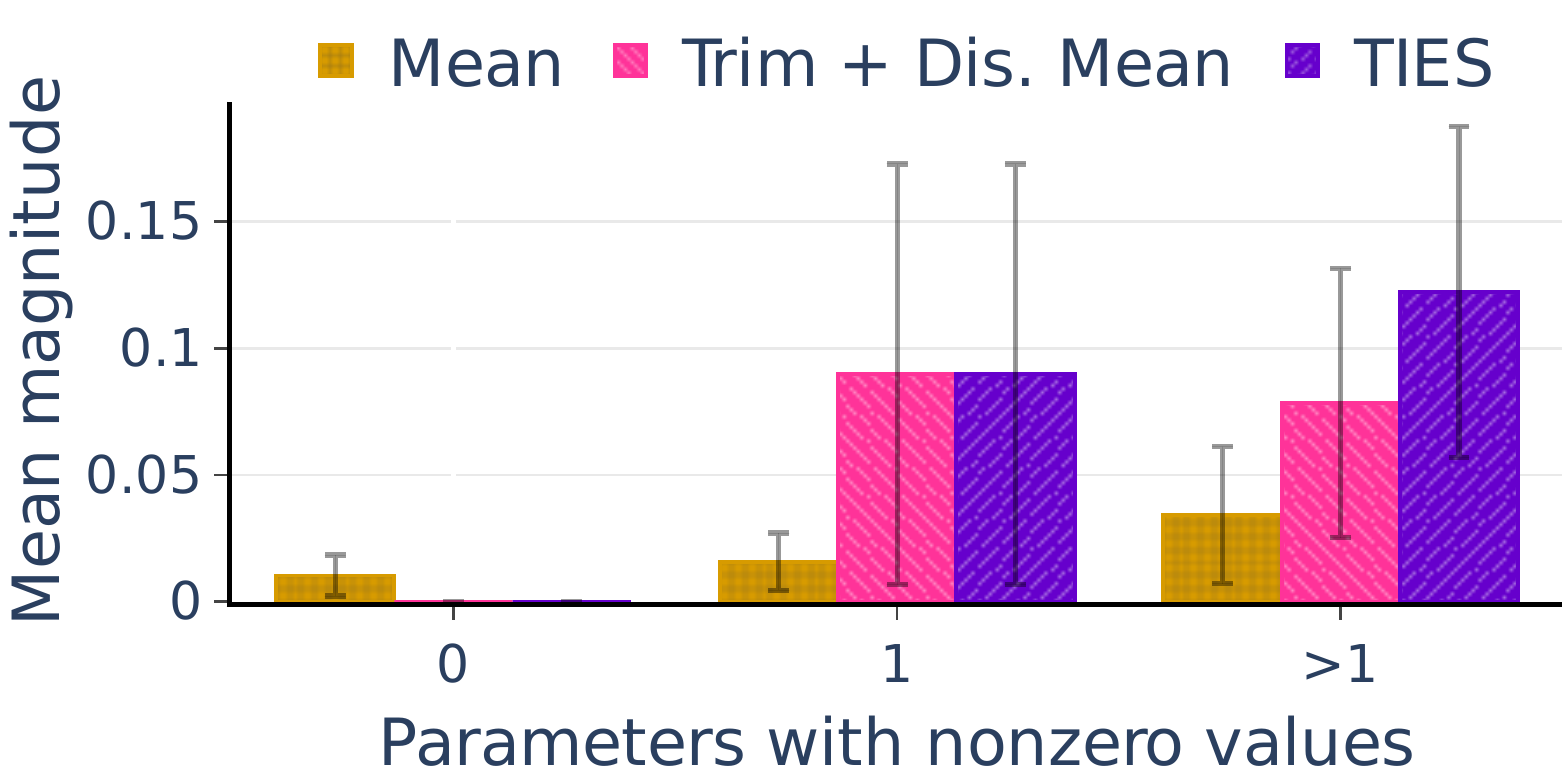}
        \caption{\label{fig:ia3_redundant_interference_full} Redundant Parameter Interference for (IA)$^3$ with STD.}
  \end{subfigure}
  \hfill
  \begin{subfigure}[b]{0.48\linewidth}
        \centering
        \includegraphics[width=\linewidth]{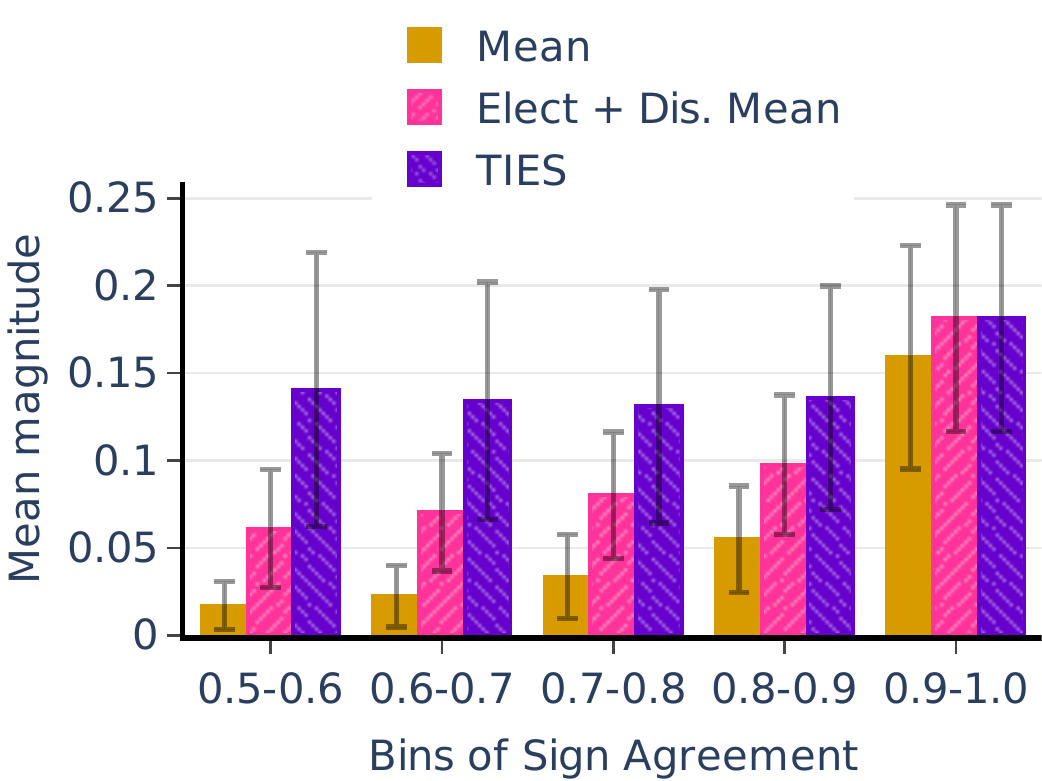}
        \caption{\label{fig:ia3_sign_interference_full} Sign Interference for (IA)$^3$ with STD.}
    
  \end{subfigure}
  \caption{
  \label{fig:app_interference_ia3} 
  Effect of different types of Merging on the Magnitudes of the Parameters. Here we additionally compare with \methodshort{} and also provide the standard deviation of parameter values. A high std implies that there is some diversity in magnitude values across different parameters.}

\end{figure}

\begin{figure}[tbh!]
  \centering
  \begin{subfigure}[b]{0.32\linewidth}
        \centering
        \includegraphics[width=\linewidth]{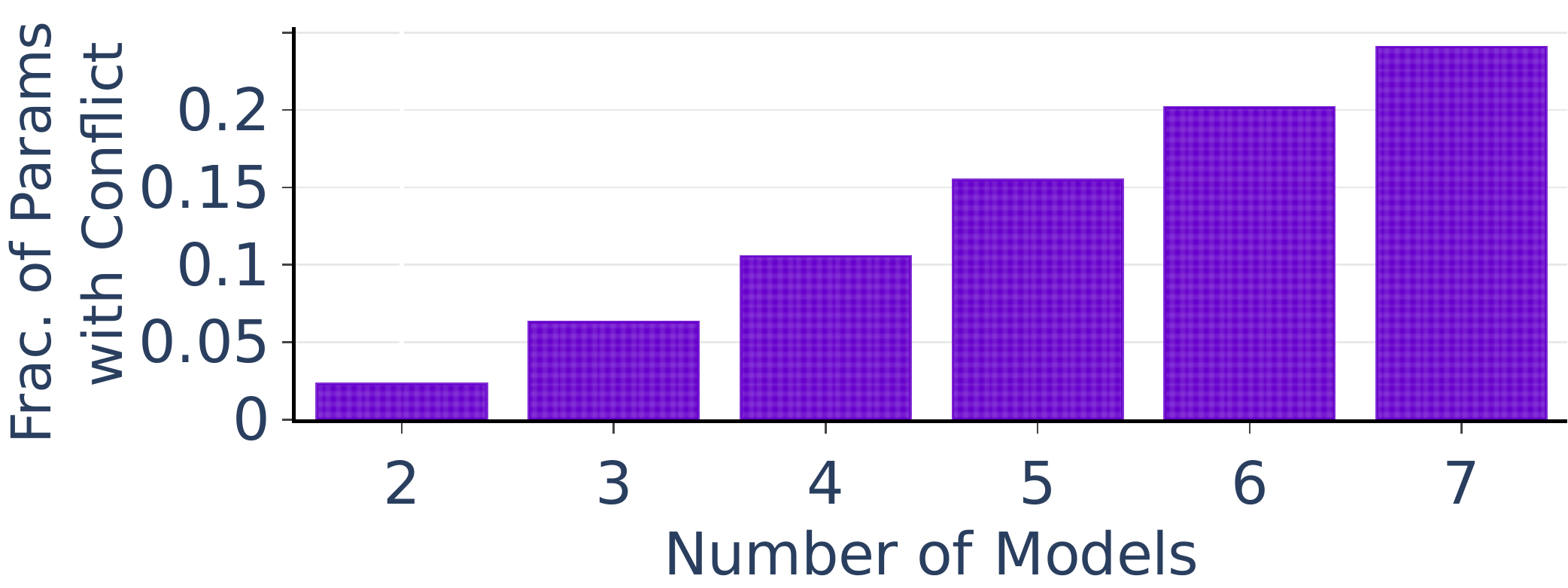}
        \caption{\label{fig:t5base_noise_scale_full} Fraction of Parameters with Sign conflicts for T5-Base model versus number of models.}
  \end{subfigure}
  \hfill
  \begin{subfigure}[b]{0.32\linewidth}
        \centering
        \includegraphics[width=\linewidth]{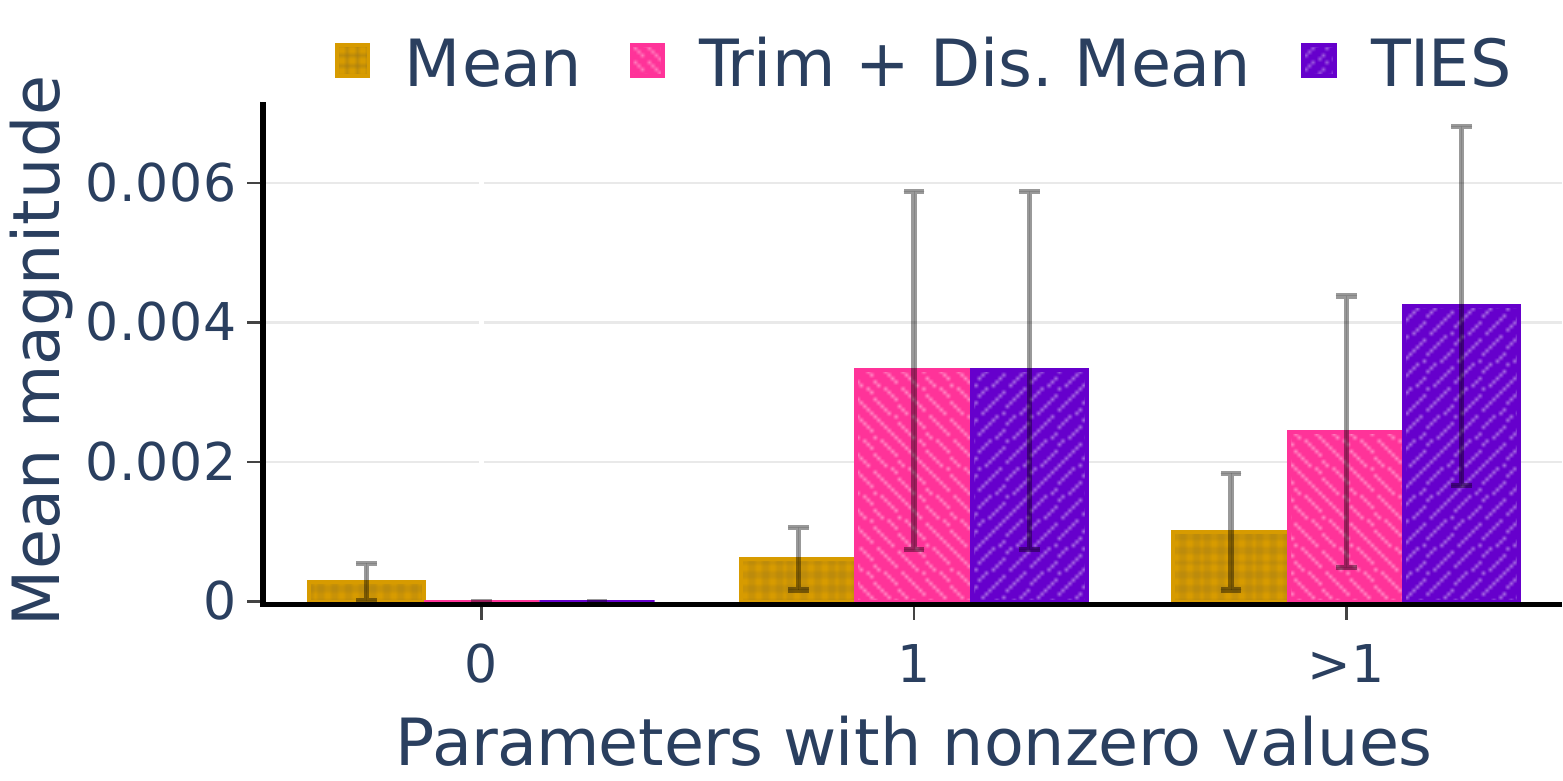}
        \caption{\label{fig:t5base_redundant_interference_full} Redundant Parameter Interference for T5-Base with STD.}
  \end{subfigure}
  \hfill
  \begin{subfigure}[b]{0.32\linewidth}
        \centering
        \includegraphics[width=\linewidth]{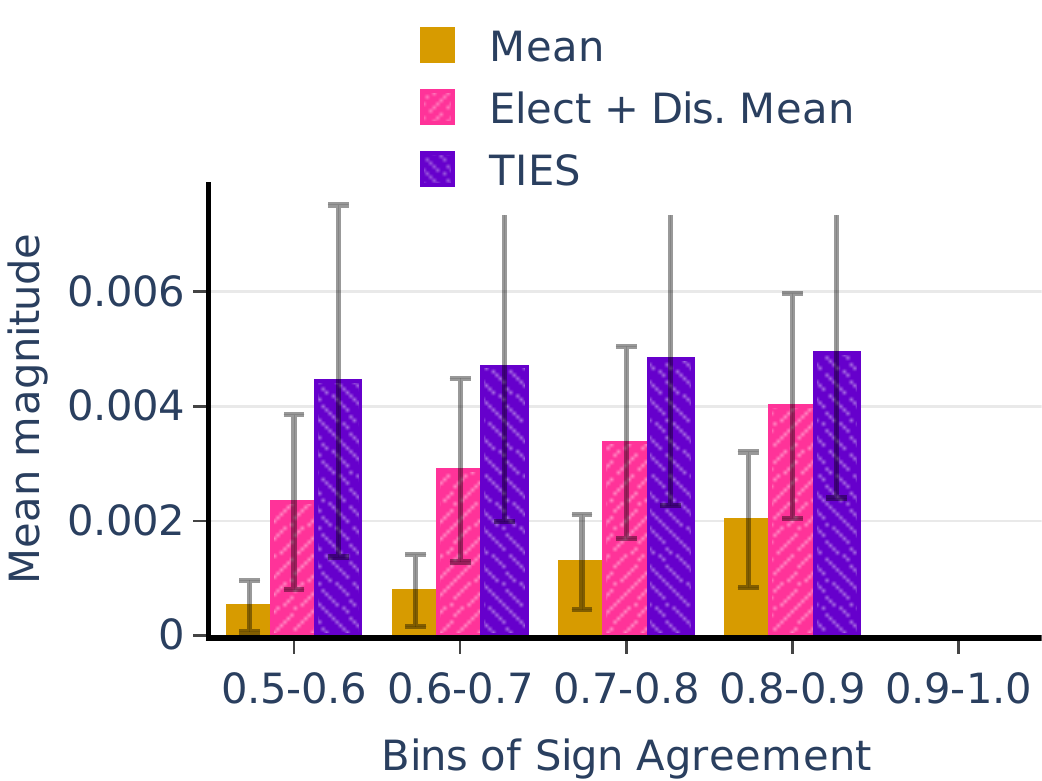}
        \caption{\label{fig:t5base_sign_interference} Sign Interference for T5-Base model with STD.}
    
  \end{subfigure}
  \caption{
  \label{fig:app_interference_t5base}
  Plots for T5-Base model.}
\end{figure}

\subsection{Detailed Results for Types of Interference and Their Effect on Merging}
\label{sec:app_effect_interference}
In Section~\ref{analysis:interference} and Figure~\ref{fig:interference}, we showed the effect of redundant parameters and sign conflicts on parameter magnitudes when comparing simple averaging vs disjoint mean after either trimming or electing and showed that performing these operations helps with the parameter magnitudes. In Figure~\ref{fig:app_interference_ia3}, we additionally compare with \methodshort{} and show that performing both trimming and electing usually results in higher magnitude and also higher standard deviation in parameter magnitudes. Higher std denotes that all parameter values in the merged model are the same and that there is a significant variation in the magnitude which is in contrast to simple averaging as it decreases the magnitude of not redundant parameters and reduces the magnitude of the influential parameters in the merged model. Similar plots for the T5-base model are provided in Figure~\ref{fig:app_interference_t5base}. 

\begin{table*}[t!]
\centering

\resizebox{\linewidth}{!}{  
\begin{tabular}{lccccccccccccc}
\toprule

\textbf{Method} & \textbf{Validation} & \textbf{Average} & \textbf{rte} & \textbf{cb} & \textbf{winogrande} & \textbf{wic} & \textbf{wsc} & \textbf{copa} & \textbf{h-swag} & \textbf{story cloze} & \textbf{anli-r1} & \textbf{anli-r2} & \textbf{anli-r3} \\
\midrule
\textbf{Zeroshot}  & - & 55.3  &  79.8  &  46.4  &  52.8  &  54.1  &  45.2  &  85  &  36.1  &  91  &  39.7  &  37.6  &  40.5 \\
\textbf{Fine-Tuned}  & - & 71.4  &  82.7  &  95.8  &  75.1  &  71.7  &  65.3  &  85.3  &  44.4  &  94.9  &  70.2  &  46.5  &  53 \\
\textbf{Multitask (All, scratch)} & - & 73.1  &  88.6  &  95.8  &  75.5  &  61.1  &  80.6  &  94.1  &  42.3  &  97.6  &  70.5  &  49.8  &  47.7 \\
\textbf{Multitask (32, scratch)}  & - & 60.9  &  74.9  &  79.2  &  59.3  &  49.2  &  63.9  &  80.9  &  39.5  &  91.6  &  49.4  &  41.9  &  40.1 \\
\textbf{Multitask (32, mean)}  & - & 65.2  &  79.8  &  83.3  &  61.6  &  54.2  &  66.7  &  85.3  &  41.1  &  94.4  &  58.1  &  46.0  &  46.5 \\
\midrule
\textbf{Averaging}  &  \redxmark & 58  &  81.2  &  58.3  &  53.8  &  55.2  &  53.5  &  80.9  &  40.1  &  92.5  &  43.3  &  39.2  &  40.2 \\
\textbf{Task Arithmetic}  & \redxmark & 59.2  &  76.5  &  79.2  &  57.7  &  51.6  &  51.4  &  66.2  &  31.4  &  81.5  &  59.8  &  47.5  &  48.2 \\
\textbf{\methodshort{}}  &  \redxmark & 64.9  &  81.2  &  87.5  &  60.8  &  59.9  &  58.3  &  80.2  &  42.6  &  91.1  &  58.1  &  46.5  &  47.4 \\
\midrule
\textbf{Fisher Merging}  & \greencmark &  62.2  &  83.3  &  83.3  &  56.7  &  54.2  &  58.3  &  83.1  &  42.2  &  94.1  &  45.9  &  41.0  &  42.2 \\
\textbf{RegMean}  & \greencmark & 58  &  81.2  &  58.3  &  53.8  &  55.2  &  53.5  &  80.9  &  40.1  &  92.5  &  43.3  &  39.2  &  40.2 \\
\textbf{Task Arithmetic}  & \greencmark & 63.9  &  74.1  &  83.3  &  62.8  &  49.1  &  49.3  &  87.5  &  41.5  &  95.3  &  60.8  &  49.4  &  50.0 \\
\textbf{\methodshort{}}  & \greencmark & 66.4  &  78.0  &  83.3  &  67.9  &  57.6  &  59.7  &  81.7  &  42.8  &  90.3  &  66.9  &  51.3  &  51.1 \\

\bottomrule
\end{tabular}
}
\caption{\label{tab:app_main_ia3} Test set performance when merging IA3 models on eleven tasks. Please refer to Section \ref{sec:main_results} for experimental details.}
\vspace{10pt}
\end{table*}

\subsection{Comprehensive Task-Level Results}
\label{sec:app_per_task_results}

We provide the task level for all the in-domain evaluation experiments in the main Table~\ref{tab:main}.
Table~\ref{tab:app_main_ia3},~\ref{tab:app_main_t5base},~\ref{tab:app_main_t5large},~\ref{tab:app_main_vitbase}, and~\ref{tab:app_main_vitlarge} provide the task level results IA3 \cite{liu2022tfew}, T5-Base, T5-Large \cite{raffel2019exploring}, ViT-B/32, and ViT-L/14 \cite{dosovitskiy2021an} respectively. The task level results of the out-of-domain experiments for T5-Base and T5-Large can be found in Table~\ref{tab:app_ood_t5base}, and \ref{tab:app_ood_t5large}. Lastly, Figure~\ref{fig:app_num_tasks_t5base}, shows the scaling of the T5-Base model as we merge different numbers of tasks.

\begin{table*}[tbh!]
\vspace{10pt}
\centering

\resizebox{\linewidth}{!}{  
\begin{tabular}{lccccccccc}
\toprule

\textbf{Method}  & \textbf{Vailidation} & \textbf{Average}  &  \textbf{paws}  &  \textbf{qasc}  &  \textbf{quartz}  &  \textbf{story\_cloze}  &  \textbf{wiki\_qa}  &  \textbf{winogrande}  &  \textbf{wsc} \\
\midrule
\textbf{Zeroshot}  & - & 53.5  &  49.9  &  35.8  &  53.3  &  48.1  &  76.2  &  50  &  61.1 \\
\textbf{Fine-tuned}  &  - & 82.8  &  94.3  &  98.3  &  80.4  &  84.7  &  95.5  &  64.1  &  62.5 \\
\textbf{Multitask}  &  - &  83.6  &  94  &  97.9  &  82.5  &  86.7  &  95  &  64.1  &  65.3 \\
\midrule
\textbf{Averaging}  & \redxmark & 65.9  &  66.4  &  82.6  &  60.2  &  49.5  &  94.1  &  50.4  &  58.3 \\
\textbf{Task Arithmetic}  & \redxmark & 73.9  &  73.3  &  93.5  &  68.2  &  76.5  &  93.7  &  55.5  &  56.9 \\
\textbf{\methodshort{}}  & \redxmark & 69.7  &  74  &  83.3  &  70.3  &  64.2  &  84.7  &  55.9  &  55.6 \\
\midrule
\textbf{Fisher Merging}  & \greencmark &  68.9  &  69.3  &  85.7  &  63.6  &  56.4  &  93.8  &  50.9  &  62.5 \\
\textbf{RegMean}  & \greencmark & 71.2  &  76.8  &  96.2  &  62.5  &  55  &  94.8  &  51.9  &  61.1 \\
\textbf{Task Arithmetic}  & \greencmark & 73.2  &  73.4  &  93.3  &  67.1  &  71.7  &  94.1  &  52.9  &  59.7 \\
\textbf{\methodshort{}}  & \greencmark & 73.9  &  79.3  &  88.6  &  71.8  &  72.9  &  82.5  &  61.3  &  61.1 \\

\bottomrule
\end{tabular}
}

\caption{\label{tab:app_main_t5base} Test set performance when merging T5-base models on seven tasks. Please refer to Section \ref{sec:main_results} for experimental details.}
\vspace{10pt}
\end{table*}

\begin{table*}[tbh!]
\centering

\resizebox{\linewidth}{!}{  
\begin{tabular}{lccccccccc}
\toprule

\textbf{Method}  & \textbf{Validation} & \textbf{Average}  &  \textbf{paws}  &  \textbf{qasc}  &  \textbf{quartz}  &  \textbf{story\_cloze}  &  \textbf{wiki\_qa}  &  \textbf{winogrande}  &  \textbf{wsc} \\
\midrule
\textbf{Zeroshot}  & - & 51.7 & 55.4 & 14.3 & 54.1 & 54.1 & 71 & 49.3 & 63.9 \\
\textbf{Fine-tuned} & - & 88.8 & 94.4 & 98.9 & 87.8 & 90.8 & 96 & 74.7 & 79.2 \\
\textbf{Multitask} & - & 88.1 & 94.2 & 98.5 & 89.3 & 92 & 95.4 & 73.5 & 73.6 \\
\midrule
\textbf{Averaging} & \redxmark  &  59.6 & 61.3 & 82.6 & 70.5 & 53.7 & 63.2 & 49.7 & 36.1 \\
\textbf{Task Arithmetic} & \redxmark & 73.5 & 79.2 & 96.8 & 80.2 & 83.6 & 58.6 & 60.2 & 55.6 \\
\textbf{\methodshort{}} & \redxmark & 74.4 & 80.5 & 96.2 & 81.8 & 78.6 & 62 & 61.9 & 59.7 \\
\midrule
\textbf{Fisher Merging} & \greencmark  & 64.6 & 60.4 & 81.7 & 75 & 60.1 & 88.6 & 50 & 36.1 \\
\textbf{RegMean} & \greencmark  & 73.2 & 86 & 96.9 & 80.7 & 78.6 & 82.6 & 51.8 & 36.1 \\
\textbf{Task Arithmetic} & \greencmark & 73.3 & 77.8 & 96 & 78.6 & 86.4 & 59.1 & 62.3 & 52.8 \\
\textbf{\methodshort{}} & \greencmark &  76.9 & 81.5 & 96.2 & 80.1 & 83.6 & 64.9 & 66.5 & 65.3 \\

\bottomrule
\end{tabular}
}
\caption{\label{tab:app_main_t5large} Test set performance when merging T5-large models on seven tasks. Please refer to Section \ref{sec:main_results} for experimental details.}
\vspace{10pt}
\end{table*}

\begin{table*}[tbh!]
\centering
\resizebox{\linewidth}{!}{  
\begin{tabular}{lcccccccccc}
\toprule

\textbf{Method}  & \textbf{Validation} & \textbf{Average}  &  \textbf{SUN397}  &  \textbf{Cars}  &  \textbf{RESISC45}  &  \textbf{EuroSAT}  &  \textbf{SVHN}  &  \textbf{GTSRB}  &  \textbf{MNIST}  &  \textbf{DTD} \\
\midrule
\textbf{Individual} & -  & 90.5  &  75.3  &  77.7  &  96.1  &  99.7  &  97.5  &  98.7  &  99.7  &  79.4 \\
\textbf{Multitask} & -  & 88.9  &  74.4  &  77.9  &  98.2  &  98.9  &  99.5  &  93.9  &  72.9  &  95.8 \\
\midrule
\textbf{Averaging} & \redxmark & 65.8  &  65.3  &  63.4  &  71.4  &  71.7  &  64.2  &  52.8  &  87.5  &  50.1 \\
\textbf{Task Arithmetic}  & \redxmark & 60.4  &  36.7  &  41  &  53.8  &  64.4  &  80.6  &  66  &  98.1  &  42.5 \\
\textbf{\methodshort{}}  & \redxmark & 72.4  &  59.8  &  58.6  &  70.7  &  79.7  &  86.2  &  72.1  &  98.3  &  54.2 \\
\midrule
\textbf{Fisher Merging}  & \greencmark & 68.3  &  68.6  &  69.2  &  70.7  &  66.4  &  72.9  &  51.1  &  87.9  &  59.9 \\
\textbf{RegMean}  & \greencmark & 71.8  &  65.3  &  63.5  &  75.6  &  78.6  &  78.1  &  67.4  &  93.7  &  52 \\
\textbf{Task Arithmetic}  & \greencmark & 70.1  &  63.8  &  62.1  &  72  &  77.6  &  74.4  &  65.1  &  94  &  52.2 \\
\textbf{\methodshort{}}  & \greencmark & 73.6  &  64.8  &  62.9  &  74.3  &  78.9  &  83.1  &  71.4  &  97.6  &  56.2 \\

\bottomrule
\end{tabular}
}
\caption{\label{tab:app_main_vitbase} Test set performance when merging ViT-B/32 models on eight tasks. Please refer to Section \ref{sec:main_results} for experimental details.}
\vspace{10pt}
\end{table*}

\begin{table*}[tbh!]
\centering

\resizebox{\linewidth}{!}{  
\begin{tabular}{lcccccccccc}
\toprule

\textbf{Method}  & \textbf{Validation} & \textbf{Average}  &  \textbf{SUN397}  &  \textbf{Cars}  &  \textbf{RESISC45}  &  \textbf{EuroSAT}  &  \textbf{SVHN}  &  \textbf{GTSRB}  &  \textbf{MNIST}  &  \textbf{DTD} \\
\midrule
\textbf{Fine-tuned} & - & 94.2  &  82.3  &  92.4  &  97.4  &  100  &  98.1  &  99.2  &  99.7  &  84.1  \\
\textbf{Multitask} & - & 93.5  &  90.6  &  84.4  &  99.2  &  99.1  &  99.6  &  96.3  &  80.8  &  97.6  \\
\midrule
\textbf{Averaging} & \redxmark & 79.6  &  72.1  &  81.6  &  82.6  &  91.9  &  78.2  &  70.7  &  97.1  &  62.8  \\
\textbf{Task Arithmetic} & \redxmark & 83.3  &  72.5  &  79.2  &  84.5  &  90.6  &  89.2  &  86.5  &  99.1  &  64.3  \\
\textbf{\methodshort{}}  & \redxmark & 86  &  76.5  &  85  &  89.3  &  95.7  &  90.3  &  83.3  &  99  &  68.8  \\
\midrule
\textbf{Fisher Merging} & \greencmark  & 82.2  &  69.2  &  88.6  &  87.5  &  93.5  &  80.6  &  74.8  &  93.3  &  70  \\
\textbf{RegMean}  & \greencmark & 83.7  &  73.3  &  81.8  &  86.1  &  97  &  88  &  84.2  &  98.5  &  60.8  \\
\textbf{Task Arithmetic}  & \greencmark & 84.5  &  74.1  &  82.1  &  86.7  &  93.8  &  87.9  &  86.8  &  98.9  &  65.6  \\
\textbf{\methodshort{}}   & \greencmark & 86  &  76.5  &  85  &  89.4  &  95.9  &  90.3  &  83.3  &  99  &  68.8  \\

\bottomrule
\end{tabular}
}
\caption{\label{tab:app_main_vitlarge} Test set performance when merging ViT-L/14 models on eight tasks. Please refer to Section \ref{sec:main_results} for experimental details.}
\vspace{10pt}
\end{table*}

\section{Implementation Details}
\label{sec:app_details}

\subsection{Compute Resources Used and Runtimes}
\label{sec:app_compute}
We executed all our experiments on Nvidia A6000 GPUs equipped with 48GB RAM. Single-task (IA)$^3$ models for eleven tasks required 1-2 hours per model, while the multitask vector took around 24 hours on four GPUs. The T5-Base and T5-Large models, based on dataset size, needed between 15 minutes and 2 hours per task, and approximately eight hours for the multitask checkpoint. Vision models ViT-B/32 and ViT-L/14 were utilized, as supplied by \citet{ilharco2023editing}.\footnote{\href{https://github.com/mlfoundations/task\_vectors\#checkpoints}{https://github.com/mlfoundations/task\_vectors\#checkpoints}}
Merge experiments were efficient, with evaluations consuming less than 2 minutes for the T5-Base, T5-Large, ViT-B/32, and ViT-L/14 experiments. 
The assessment of (IA)$^3$ models, due to the necessity of using multiple templates from prompt sources and median result calculations across all templates, required approximately one hour per 11 dataset evaluation. 

\begin{table*}[tbh!]
\vspace{10pt}
\centering
\begin{tabular}{lccccccc}
\toprule

\textbf{Model}  &  \textbf{Average}  &  \textbf{cosmos\_qa}  &  \textbf{social\_iqa}  &  \textbf{quail}  &  \textbf{wic}  &  \textbf{copa}  &  \textbf{h-swag}  \\
\midrule
\textbf{PAWS}  &  35.9  &  18.8  &  25  &  24.8  &  68.8  &  56.2  &  21.9  \\
\textbf{QASC}  &  34.9  &  15.6  &  21.9  &  25.1  &  75  &  53.1  &  18.8  \\
\textbf{QUARTZ}  &  37.4  &  31.2  &  18.8  &  24.3  &  71.9  &  59.4  &  18.8  \\
\textbf{Story Cloze}  &  35  &  6.2  &  25  &  25.6  &  75  &  65.6  &  12.5  \\
\textbf{Wiki QA}  &  24.5  &  18.8  &  21.9  &  24.9  &  28.1  &  43.8  &  9.4  \\
\textbf{Winogrande}  &  28.3  &  18.8  &  25  &  25.7  &  34.4  &  43.8  &  21.9  \\
\textbf{WSC}  &  31.7  &  21.9  &  21.9  &  24.6  &  62.5  &  46.9  &  12.5  \\
\midrule
\textbf{Pretrained}  &  31.1  &  21.9  &  18.8  &  24.1  &  65.6  &  43.8  &  12.5  \\
\textbf{Averaging}  &  31.7  &  21.9  &  21.9  &  24.6  &  68.8  &  37.5  &  15.6  \\
\textbf{Fisher Merging}  &  33.8  &  15.6  &  21.9  &  24.9  &  65.6  &  53.1  &  21.9  \\
\textbf{Task Arithmetic}  &  31.9  &  15.6  &  31.2  &  25.7  &  28.1  &  68.8  &  21.9  \\
\textbf{RegMean}  &  34.3  &  23.1  &  28.1  &  24.9  &  48.4  &  62.5  &  18.8  \\
\textbf{\methodshort{}}  &  35.3  &  21.9  &  25  &  25.7  &  50  &  65.6  &  23.8  \\

\bottomrule
\end{tabular}
\caption{\label{tab:app_ood_t5base}  Out-of-Distributon performance of T5-Base model checkpoints on six tasks. Please refer to Section \ref{sec:main_results} for experimental details.}
\end{table*}

\subsection{Employed Datasets and Associated Licences}
\label{sec:app_dataset}
We use the following datasets in the paper with the following licenses. 
ANLI \citep{nie2019adversarial}, WiC \citep{pilehvar2018wic}, WSC \citep{wsc}, and Story Cloze \citep{sharma2018tackling}, QuaRTz \cite{tafjord2019quartz}, Cars \citep{cars}, GTSRB \citep{gtsrb} are under Creative Commons License. Winogrande \citep{sakaguchi2020winogrande}, QASC \cite{khot2020qasc} are under Apache license. COPA \citep{copa} is under a BSD-2 Clause license. H-SWAG \citep{zellers2019hellaswag}, EuroSAT \citep{eurosat}, is under MIT Licence. MNIST \citep{lecun1998mnist} is under Gnu General Public License. We could not find the licences of DTD \citep{dtd}, RESISC45 \citep{cheng2017remote}, SUN397 \citep{sun397}, SVHN \citep{svhn}, CB \citep{cb}, RTE \citep{dagan2005pascal}), and PAWS \cite{paws2019naacl} but they are publically for research use.

\subsection{Details of the Motivation Experiments}
\label{sec:app_motivation}

For both Figure~\ref{fig:reset-bottomk},~and~ \ref{fig:imp_conflict} in Section~\ref{sec:background}, we perform experiment on the eleven (IA)$^3$ models used in our PEFT merging experiments (\S~\ref{sec:main_results}). For a Figure similar to Fig.~\ref{fig:imp_conflict} demonstrating the fraction of parameters with a sign conflict for T5-base model, please refer to Fig.~\ref{fig:t5base_noise_scale_full}.

\begin{table*}[tbh!]
\vspace{10pt}
\centering
\begin{tabular}{lccccccc}
\toprule

\textbf{Model}  &  \textbf{Average}  &  \textbf{cosmos\_qa}  &  \textbf{social\_iqa}  &  \textbf{quail}  &  \textbf{wic}  &  \textbf{copa}  &  \textbf{h-swag}  \\
\midrule
\textbf{PAWS}  &  32.3  &  25  &  28.1  &  25.6  &  56.2  &  46.9  &  12.5  \\
\textbf{QASC}  &  33.4  &  21.9  &  28.1  &  25.5  &  43.8  &  62.5  &  18.8  \\
\textbf{QUARTZ}  &  28.7  &  25  &  25  &  25.1  &  25  &  53.1  &  18.8  \\
\textbf{Story Cloze}  &  32.1  &  21.9  &  34.4  &  26.8  &  46.9  &  53.1  &  9.4  \\
\textbf{Wiki QA}  &  27.1  &  25  &  28.1  &  25.2  &  28.1  &  46.9  &  9.4  \\
\textbf{Winogrande}  &  32.4  &  31.2  &  18.8  &  25.6  &  43.8  &  62.5  &  12.5  \\
\textbf{WSC}  &  29.7  &  25  &  25  &  25.1  &  37.5  &  56.2  &  9.4  \\
\midrule
\textbf{Pretrained}  &  27.6  &  21.9  &  21.9  &  24.9  &  28.1  &  56.2  &  12.5  \\
\textbf{Averaging}  &  30.4  &  31.2  &  25  &  26.3  &  31.2  &  59.4  &  9.4  \\
\textbf{Fisher Merging}  &  32  &  34.4  &  25  &  26.1  &  40.6  &  56.2  &  9.4  \\
\textbf{Task Arithmetic}  &  33.3  &  21.9  &  34.4  &  24.6  &  40.6  &  59.4  &  18.8  \\
\textbf{RegMean}  &  36  &  34.4  &  28.1  &  25.3  &  62.5  &  50  &  15.6  \\
\textbf{\methodshort{}}  &  40.4  &  31.2  &  43.8  &  26.6  &  59.4  &  59.4  &  21.9  \\

\bottomrule
\end{tabular}
\caption{\label{tab:app_ood_t5large} Out-of-Distributon performance of T5-Large model checkpoints on six tasks. Please refer to Section \ref{sec:main_results} for experimental details.}
\end{table*}

\subsection{Merging in the absence of the Validation Set}
\label{sec:app_validation}

In our investigation into scenarios where a validation set is not available, we have devised a recipe and identified the optimal hyperparameters, employing the PEFT experimental procedure detailed in Section \ref{sec:main_results}. This approach was applied to the eleven task-specific models presented in the same section, utilizing the \methodshort{} method for tuning the hyperparameters. Our preliminary estimates for the hyperparameters were $k = 20$ and $\lambda$ close to 1.
The hyperparameter search was conducted using the eleven task-specific (IA)$^3$ models, with $k \in \{10, 20, 30\}$, and $\lambda$ spanning from 0.8 to 3.0, in increments of 0.1. The results of this comprehensive search indicated an optimal value of $k = 20$, with values of $\lambda = 0.9$, $\lambda = 1.0$, and $\lambda = 1.1$ demonstrating equivalent performance. To maintain simplicity in our model, we chose a $\lambda$ value of 1.
Thus, the final selection of parameters for \methodshort{} is $k=20$, signs based on mass, the disjoint mean, and a $\lambda$ value of 1.

\begin{figure}
    \centering
    \includegraphics[width=0.5\linewidth]{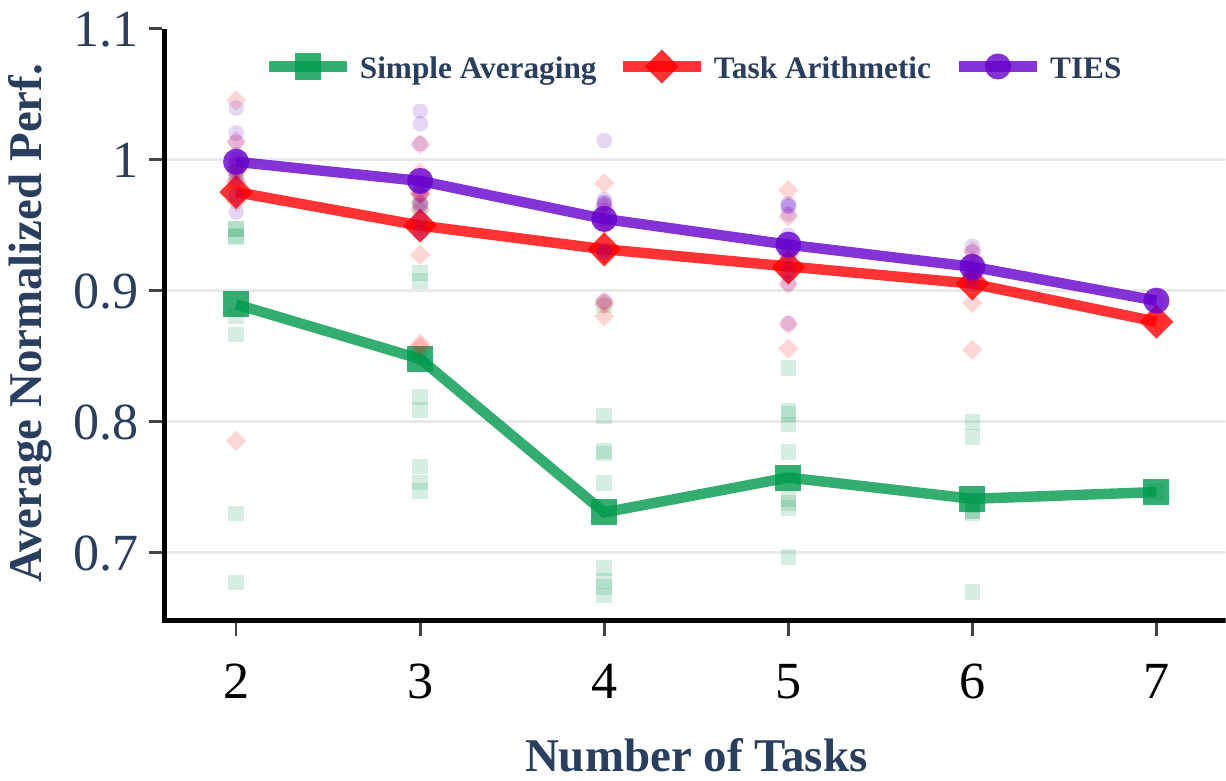}
    \captionsetup{type=figure}
    \caption{\label{fig:app_num_tasks_t5base} T5-Base with increasing number of task being merged. Average performance when merging a different number of tasks.}
\end{figure}

\subsection{Merging Different Number of Tasks}
\label{sec:app_num_tasks}
Here we provide some additional details on the experiments when merging different numbers of tasks. In Fig.~\ref{fig:num_tasks}, we perform the experiment with T5-Large when merging the seven tasks considered in Tab.~\ref{tab:main} and described in Sec.~\ref{sec:main_results}. The x-axis shows the different number of tasks being merged. Note that when merging $T$ tasks, we have a total of $\binom{7}{T}$ combinations. However, in our experiment, we sample at most 10 distinct combinations for each value of $T$. A similar plot for the T5-Base model is shown in Fig.~\ref{fig:app_num_tasks_t5base}.

In Figure~\ref{fig:num_tasks}, for each number of tasks we take at most $10$ random subsets of the 8 tasks we were considering. The solid line is the average of the merged model's performance from these different runs. Below we provide the optimal $\lambda$ values for the different subsets of tasks we merged for both \methodshort{} and Task Arithmetic, note that for averaging the $\lambda = \frac{1}{\# tasks}$ always. Each entry in the list is the optimal $\lambda$ for a particular subset of tasks selected on the validation set.

(2 tasks) TIES $\rightarrow$ [1.7, 1.9, 2, 2, 1.1, 1.5, 1.6, 1.8, 1.9, 1., 5]\\ 
(2 tasks) Task Arithmetic $\rightarrow$ [1, 0.9, 1, 1, 0.9, 1, 0.9, 0.9, 0.9, 1]

(3 tasks) TIES $\rightarrow$ [1.2, 2, 1.5, 1.9, 1.8, 1.7, 1.4, 2, 3, 1.9]\\
(3 tasks) Task Arithmetic $\rightarrow$ [1, 0.7, 0.7, 1, 1, 0.9, 0.7, 0.7, 0.9, 1]

(4 tasks) TIES $\rightarrow$ [1.5, 1.3, 1.3, 1.8, 2.3, 1.7, 1.8, 1.7, 1.9, 1.5]\\
(4 tasks) Task Arithmetic $\rightarrow$ [0.8, 0.7, 0.7, 0.7, 0.6, 0.7, 0.7, 0.8, 0.6, 0.7]

(5 tasks) TIES $\rightarrow$ [2, 2, 2, 1.8, 1.7, 2, 1.6, 2.1, 1.6, 1.3]\\
(5 tasks)Task Arithmetic $\rightarrow$ [0.7, 0.8, 0.6, 0.8, 0.7, 0.6, 0.6, 0.6, 0.6, 0.7]

(6 tasks) TIES $\rightarrow$ [1.6, 1.7, 1.7, 1.2, 1.7, 1.7, 1.5]\\
(6 tasks) Task Arithmetic $\rightarrow$ [0.6, 0.5, 0.5, 0.5, 0.7, 0.5, 0.6]

(7 tasks) TIES $\rightarrow$ [1.7]\\
(7 tasks) Task Arithmetic $\rightarrow$ [0.5]

\subsection{Training Details}
\label{sec:app_training_details}

In our research, we utilized two variants of the T5 model, specifically the T5-base and T5-large models, which were trained to a maximum of 75,000 steps. An effective training batch size of 1024 was implemented, alongside a learning rate (lr) of 0.0001. We instituted an early stopping mechanism with a patience threshold of 5 to prevent overfitting. During the training process, bfloat16 was adopted to curtail GPU memory expenditure, and the maximum sequence length was set at 128. In contrast, for the PEFT configuration of the (IA)$^3$ approach on the T0-3B model, we modified our parameters. An effective training batch size of 16 was deployed along with an evaluation batch size of 32, while maintaining the learning rate at 0.0001. To accommodate the model's complexity, the early stopping patience was augmented to 10. We do not use any lr scheduler and weight decay for any of our model training.

For the purpose of evaluation, we perform \textit{rank classification}. In this method, the model's log probabilities for all potential label strings are ranked. The model's prediction is deemed accurate if the choice ranked highest aligns with the correct answer. It should be noted that rank classification evaluation can accommodate both classification tasks and multiple-choice tasks.

\end{document}